\documentclass[format=acmsmall, noacm]{acmart}

\usepackage{amsmath}
\usepackage{array}
\usepackage{balance}
\usepackage{booktabs} 
\usepackage{graphicx}
\usepackage{multirow}
\usepackage{rotating}
\usepackage{soul}
\usepackage{url}
\usepackage[utf8]{inputenc}
\usepackage{subcaption}
\usepackage[export]{adjustbox}
\usepackage{stackengine}
\usepackage{amsfonts}
\usepackage{tikz}
\usepackage{amsthm}
\usepackage[normalem]{ulem}
\useunder{\uline}{\ul}{}
\usepackage[right]{lineno}

\theoremstyle{definition}

\newif\ifdraft
\drafttrue

\newcommand{\dv}{DVs}
\newcommand{\diffvectors}{Diff-Vectors}

\newcommand{\side}[1]{\begin{sideways}{#1}\end{sideways}}

\begin{document}

%
\title[Diff-Vectors for Authorship Analysis]{Same or Different?
Diff-Vectors for Authorship Analysis}


\author{Silvia Corbara} \email{silvia.corbara@sns.it}
\orcid{0000-0002-5284-1771} \affiliation{%
\institution{Scuola Normale Superiore, 56126 Pisa, Italy.}
}

\author{Alejandro Moreo} \email{alejandro.moreo@isti.cnr.it}
\orcid{0000-0002-0377-1025} \affiliation{%
\institution{Istituto di Scienza e Tecnologie dell'Informazione,
Consiglio Nazionale delle Ricerche, 56124 Pisa, Italy.}
}

\author{Fabrizio Sebastiani} \email{fabrizio.sebastiani@isti.cnr.it}
\orcid{0000-0003-4221-6427} \affiliation{%
\institution{Istituto di Scienza e Tecnologie dell'Informazione,
Consiglio Nazionale delle Ricerche, 56124 Pisa, Italy.}
}

\renewcommand{\shortauthors}{Corbara, Moreo, Sebastiani}

\begin{abstract}
  \noindent In this paper we investigate the effects on authorship
  identification tasks (including authorship verification, closed-set
  authorship attribution, and closed-set and open-set same-author
  verification) of a fundamental shift in how to conceive the
  vectorial representations of documents that are given as input to a
  supervised learner. In ``classic'' authorship analysis a feature
  vector represents a document, the value of a feature represents (an
  increasing function of) the relative frequency of the feature
  in the document, and the class label represents the author of the
  document. We instead investigate the situation in which a feature
  vector represents an unordered \emph{pair} of documents, the value
  of a feature represents the absolute difference in the relative
  frequencies (or increasing functions thereof) of the feature in the
  two documents, and the class label indicates whether the two
  documents are from the same author or not. This latter
  (learner-independent) type of representation has been occasionally
  used before, but has never been studied systematically. We argue
  that it is advantageous, and that in some cases (e.g., authorship
  verification) it provides a much larger quantity of information to
  the training process than the standard representation. The
  experiments that we carry out on several publicly available datasets
  (among which one that we here make available for the first time)
  show that feature vectors representing pairs of documents (that we
  here call \emph{\diffvectors}) bring about systematic improvements
  in the effectiveness of authorship identification tasks, and
  especially so when training data are scarce (as it is often the case
  in real-life authorship identification scenarios). Our experiments
  tackle same-author verification, authorship verification, and
  closed-set authorship attribution;
  while DVs are naturally geared for solving the 1st, we also provide
  two novel methods for solving the 2nd and 3rd that use a solver for
  the 1st as a building block. The code to reproduce our experiments
  is open-source and available
  online.\footnote{\url{https://github.com/AlexMoreo/diff-vectors}}
 
\end{abstract}


\maketitle



\section{Introduction}
\label{sec:introduction}

\noindent Recent years have seen an increased interest in automated
\emph{authorship analysis}, a set of tasks aiming to infer the
characteristics of the author of a text of unknown or disputed
paternity. Authorship analysis is concerned with inferring
characteristics such as the gender \cite{Koppel02}, the age group
\cite{Gollub:2013ri}, or the native language \cite{Tetreault:2012fu}
of the author, among others; these subtasks usually go under the name
of \emph{author profiling} \cite{Argamon:2009pw}. Alternatively,
authorship analysis may be concerned with inferring the
\emph{identity} of the author; tasks in which this is the goal are
collectively referred to as \emph{authorship identification} tasks,
and include \emph{authorship verification} (AV -- the task of
predicting whether a given author is or not the author of a given
anonymous text \cite{Stamatatos:2016ij}), \emph{authorship
attribution} (AA -- the task of predicting who among a given set of
candidates is the most likely author of a given anonymous text
\cite{Juola:2006fk, Koppel:2009ix, Stamatatos2009:yq}), and
\emph{same-author verification} (SAV -- the task of predicting whether
two given documents are by the same, possibly unknown, author or not
\cite{Koppel:2014bq}). Authorship analysis has several applications,
e.g., in supporting the work of philologists who try to identify the
authors of texts of literary or historical value
\cite{Benedetto:2013mb, Corbara:2019cq, Kabala:2020bu,
Kestemont:2015lp, Mosteller:1964gb, Savoy:2019qr, Tuccinardi:2017yg},
or in aiding linguistic forensics experts in crime prevention or
criminal investigation \cite{Chaski:2005pd, Larner:2014kl,
Rocha:2017yy}.

All of these tasks are usually approached as \emph{text
classification} tasks, whereby a supervised machine learning
algorithm, using a set of labelled documents, is used to train a
classifier to perform the required prediction task. As in many
supervised learning endeavours, each training example is usually
represented as a vector of features, where the value of a feature in a
vector usually corresponds to the relative frequency with which a
certain linguistic phenomenon (say, an exclamation mark, or a
POS-gram) occurs within the document.


In this paper we carry out an in-depth analysis of an alternative
method for generating vectorial representations of texts for
authorship identification. Specifically, while in the standard
representation methodology a vector represents a document, in this
alternative method a vector represents an unordered \emph{pair} of
different documents. While in the standard methodology the value of a
feature is (an increasing function of) the relative frequency of
occurrence of a given linguistic phenomenon in the document,
in this alternative method it is the absolute value of the
\emph{difference} between the relative frequencies (or increasing
functions thereof) of this phenomenon in the two documents. Since
these vectors represent differences, we call these representations
\emph{\diffvectors} (DVs). While in the standard methodology the class
label is the author of the document, in this DV-based methodology the
class label is one of the two classes \textsf{Same} or
\textsf{Different} (standing for ``same author'' or ``different
authors'', respectively).

Technically, this latter type of representation is not novel, since it
was first described (to the best of our knowledge) by
\citet{Koppel:2014bq}. However, curiously enough, the goal of
\cite{Koppel:2014bq} was to propose a different method (the
``impostors'' method for SAV), and its authors mention the DV-based
representation only to dismiss it as a ``simplistic baseline method''
\cite[p.\ 179]{Koppel:2014bq}. Since then, the use of DVs has never
been studied systematically; to carry out such a systematic study is
the goal of the present paper.

We carry out extensive experiments on a number of publicly available
datasets (among which one that we here make available for the first
time) representative of different textual genres, lengths, and
styles. In these experiments we tackle different authorship
identification tasks, including SAV (for which DVs are naturally
geared), AA, and AV; for these two latter tasks we propose two new
methods, \emph{Lazy AA} and \emph{Stacked AA} (two AA methods that can
also be used for AV) that solve AA by using a DV-based SAV classifier
as a building block. Our experiments show that the DV-based
representation is advantageous, since it brings about substantially
increased effectiveness at the price of a tolerable increase in
computational cost. The experiments also show that DVs bring about
substantial improvements especially in low-resource authorship
analysis tasks, i.e., in tasks characterised by small quantities of
training data (which is the case in many real-life authorship analysis
scenarios, such as those dealing with ancient texts). Like the
standard representation, the DV-based representation is
learner-independent, i.e., it can be used in connection with any
(supervised or unsupervised) learning method.

The rest of the paper is structured as follows. In
Section~\ref{sec:dvs} we formally describe DVs and justify why they
look like a superior means of representing authorship-related
information. In Section~\ref{sec:recasting} we describe algorithms for
casting authorship identification tasks (such as AV or AA) in terms of
SAV (the task that DVs are naturally designed
for). Section~\ref{sec:experiments} reports the results of our
experiments; in particular, Section~\ref{sec:intrinsic} discusses our
``intrinsic'' evaluation of \dv, i.e., one in terms of same-author
verification, while Section~\ref{sec:extrinsic} discusses an
``extrinsic'' evaluation of DVs, i.e., one in terms of downstream
tasks such as AV and AA.
Section~\ref{sec:related} discusses related work, while
Section~\ref{sec:conclusion} wraps up, also pointing at avenues for
further research.



\section{\diffvectors\ for authorship identification}
\label{sec:dvs}



\subsection{Authorship identification tasks}
\label{sec:authorshipanalysis}

\noindent We assume a finite set $\mathcal{A}$ of authors (where
$\mathcal{A}$ will be often called the \emph{codeframe}) and a domain
$\mathcal{D}$ of documents. For each document $x_{i}\in \mathcal{D}$
we indicate by $y_{i}\in\mathcal{A}$ the true author of $x_{i}$. We
also assume the existence of a training set
$\mathcal{L}=\{(x_{i},y_{i})\}_{i=1}^{n}$ of documents of known
paternity.

We define \emph{authorship verification} (AV) as the task of
predicting, given a document $x_{i}$ and a candidate author
$A^{*}\in\mathcal{A}=\{A_{1}, \ldots, A_{m}\}$, whether $A^{*}$ is the
author of $x_{i}$ or not, where the labels $y_{1}, \ldots, y_{n}$ of
the training documents are in $\mathcal{A}=\{A_{1}, \ldots, A_{m}\}$,
with $m\geq2$.\footnote{Note that, at training time, we assume to know
the paternity (i.e., the labels) of documents written by authors other
than $A^{*}$. Alternatively, AV can be formulated as a problem in
which $m=2$ and $\mathcal{A}=\{A^{*}, \overline{A^{*}}\}$, in which
class $\overline{A^{*}}$ collectively represents the production of
authors other than $A^{*}$. This special case will be discussed more
in detail in Section~\ref{sec:nativeav}.
}

We define \emph{(closed-set) authorship attribution} (AA) as the task
of predicting, given a document $x_{i}$ and $m$ candidate authors
$\mathcal{A}=\{A_{1}, \ldots, A_{m}\}$, (one of whom is assumed to be
the author of $x_{i}$), who among the members of $\mathcal{A}$ is
the 
author of $x_{i}$, where the labels of the training documents are in
$\mathcal{A}=\{A_{1}, \ldots, A_{m}\}$, with $m\geq2$.\footnote{In
real cases we may not be certain that the author of $x_{i}$ is indeed
in $\mathcal{A}=\{A_{1}, \ldots, A_{m}\}$; in these cases, closed-set
AA amounts to indicating who, among the authors in
$\mathcal{A}=\{A_{1}, \ldots, A_{m}\}$, is the \emph{most likely}
author of $x_{i}$.}

We define \emph{same-author verification} (SAV) as the task of
predicting, given two unlabelled documents $x_{i}$ and $x_{j}$, if
they are by the same author or not, where the labels
of the training documents are in
$\mathcal{A}=\{A_{1}, \ldots, A_{m}\}$, with $m\geq 2$. This task
admits two different variants, i.e., (i) \emph{closed-set} SAV, which
corresponds to the setup in which the authors of the unlabelled
documents are assumed to be in $\mathcal{A}$, and (ii) \emph{open-set}
SAV, where the authors of the unlabelled documents are not necessarily
in $\mathcal{A}$.

Note that terminology is somehow variable across the authorship
analysis literature, and some of the above tasks may be defined
slightly differently in other works. For instance, authorship
verification is sometimes defined (see e.g., \citep{Kestemont:2021eu})
as the task of predicting whether, given a document $x_{i}$ and one or
more documents known to be by a candidate author $A^{*}$, also $x_{i}$
is by $A^{*}$. In this latter definition authorship verification
shares some characteristics with ``our'' AV (in the fact that a
candidate author $A^{*}$ for document $x_{i}$ is considered) and with
``our'' SAV (in the fact that we check whether $x_{i}$ is by $A^{*}$
by testing if $x_{i}$ is by the same author as other texts known to be
by $A^{*}$). Our definition of AV and SAV are, we think, cleaner,
since they clearly separate (i) the task of predicting whether a
document $x_{i}$ is by a candidate author $A^{*}$, from (ii) the task
of predicting whether a document $x_{i}$ is by the same author as some
other document. Our definitions are also more general, since ``our''
SAV does not assume the author of one of the two documents to be
known.

\subsection{\diffvectors}
\label{sec:diff}

\noindent In ``standard'' authorship identification, each document
$x_{i}$ is represented via a labelled vector $\mathbf{x}_{i}$ of
features, where each feature usually represents a linguistic
phenomenon that may occur (possibly several times) in a document of
$\mathcal{D}$, the label $y_{i}\in \mathcal{A}$ represents the true
author of $x_{i}$, and the value $\mathbf{x}_{i}^{k}$ of the $k$-th
feature in vector $\mathbf{x}_{i}$ represents a non-decreasing
function (e.g., tfidf) of the relative frequency of the linguistic
phenomenon in $x_{i}$.
For instance, if the $k$-th feature stands for character 3-gram
``car'', then the value of $\mathbf{x}_{i}^{k}$ may be the number of
occurrences of character 3-gram ``car'' in $x_{i}$ divided by the
number of all character 3-grams that $x_{i}$ contains.


We here study an alternative type of vectorial representation for
authorship identification tasks. Here, a labelled vector
$\mathbf{x}_{ij}$ represents an \emph{unordered pair} $(x_{i},x_{j})$
of documents in $\mathcal{D}$ such that $i\not = j$, each feature
represents a linguistic phenomenon that may occur (possibly several
times) in a document of $\mathcal{D}$, the label
$y_{ij}\in \mathcal{P}=\{\textsf{Same},\textsf{Different}\}$ indicates
whether the true authors of $x_{i}$ and $x_{j}$ are the same person or
not, and the value $\mathbf{x}_{ij}^{k}$ of the $k$-th feature in
vector $\mathbf{x}_{ij}$ represents the absolute difference between
non-decreasing functions of the relative frequencies of the linguistic
phenomenon in $x_{i}$ and $x_{j}$. (In this section we provisionally
assume this function to be the identity function $f(x)=x$, while in
the sections to come this function will be some well-established
feature weighting function.)
Since the \emph{difference} between relative frequencies is central to
the definition of these vectors, we call them \emph{\diffvectors}
(DVs).

If we have chosen our features well, i.e., if the frequencies of
occurrence of the corresponding linguistic phenomena are indeed
indicative of authorship, when two documents have been written by the
same author the values $\mathbf{x}_{ij}^{k}$ of these features will be
low, since the above frequencies will be similar in the two
documents. In other words, DVs belonging to class \textsf{Same} will
tend to be characterised by low feature values and low norms, while
vectors belonging to class \textsf{Different} will tend to be
characterised by high feature values and high norms. The
quintessential example of a DV likely to be in class \textsf{Same} is
the vector of all 0's, since the fact that for all features the
frequency of occurrence of the feature in the two documents is
identical, is highly indicative of the fact that (if the features have
been chosen well) the two authors are the same person. Conversely, the
quintessential example of a DV likely to be in class
\textsf{Different} is (if feature values are all normalised) a vector
of all 1's, since it represents two documents with maximally different
frequencies of occurrences for all features. All DVs fall, if
normalised, in the unit hypercube.

\begin{figure}[t]
  \begin{center}
    \scalebox{.55}[.55]{\includegraphics{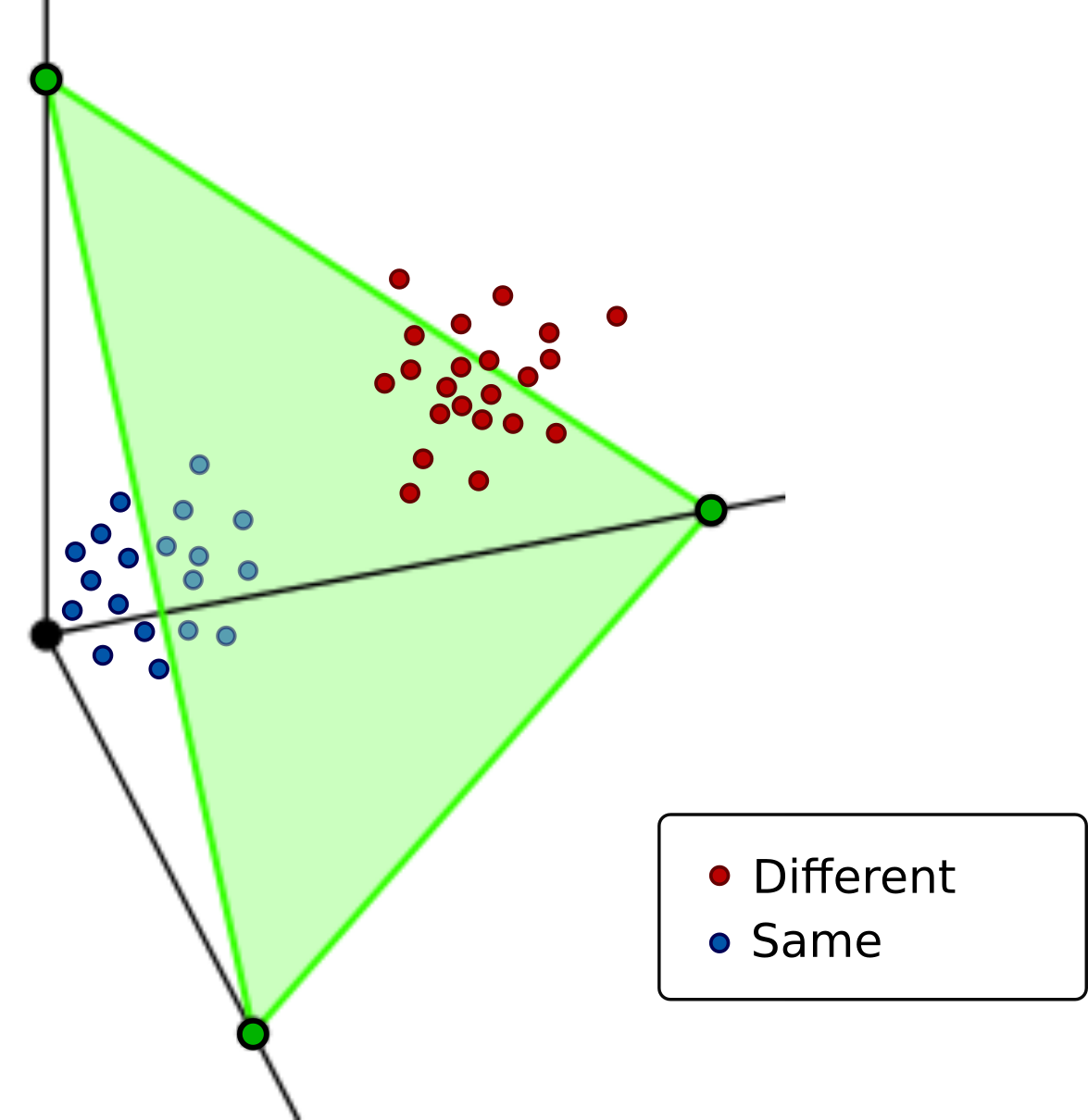}} \\
    \caption{3-dimensional example of the surface (in green) that
    (ideally) separates the region of DVs belonging to \textsf{Same}
    (which corresponds to the tetrahedron comprised between the
    separating surface and the origin of the axes) and the region of
    DVs belonging to \textsf{Different}, in the linear case. When the
    number of features is $t$, the tetrahedron becomes a $t$-simplex
    and the separating surface is a $(t-1)$-simplex.}
    \label{fig:ksimplex}
  \end{center}
\end{figure}

More in general, if a DV belongs to class \textsf{Same}, DVs that lie
between it and the vector of all 0's will also tend (if we have chosen
our features well) to belong to \textsf{Same}. As a result, the region
that contains the DVs belonging to \textsf{Same} will tend to be the
portion falling in the non-negative orthant of a star-convex region
centred at the origin of the axis.\footnote{The \emph{non-negative
orthant} is the generalisation to $t>2$ dimensions of the 1st quadrant
of the familiar 2-dimensional Cartesian space. A \emph{star-convex
region} centred at point $\mathbf{x}_{0}$ is a region of
$t$-dimensional space in which for every point $\mathbf{x}$ in the
region all points between $\mathbf{x}_{0}$ and $\mathbf{x}$ are also
in the region.} In particular, if \textsf{Same} and \textsf{Different}
are linearly separable, and if $t$ is the dimensionality of the
feature space, the region that contains all the DVs belonging to
\textsf{Same} will tend to be (see Figure~\ref{fig:ksimplex}) a
$t$-simplex (in $t=3$ dimensions: a tetrahedron) with an orthogonal
corner, and the separating surface will tend to be a $(t-1)$-simplex
(in $t=3$ dimensions: a triangle).\footnote{A \emph{$t$-simplex} is
the generalisation to $t>3$ dimensions of the 2-dimensional notion of
triangle and the 3-dimensional notion of tetrahedron. A $t$-simplex
with an \emph{orthogonal corner} is one that has a vertex such that
all its adjacent edges are pairwise orthogonal. }
%


Any set of labelled documents $\mathcal{L}=\{(x_{1},y_{1})$, \ldots,
$(x_{n},y_{n})\}$ can be represented either in the standard way or via
DVs. One of the main differences between the two representations is
that the ``standard'' representation gives rise to $n$ labelled
vectors, while the alternative representation gives rise to $n(n-1)/2$
labelled vectors.
The other main difference is that a classifier using the ``standard''
representation attempts to predict, given an unlabelled document, its
true author, while a classifier using the DV-based representation
attempts to predict, given two unlabelled documents, whether the two
documents are or not by the same author. \emph{In other words, the
standard representation is geared towards AV or AA, while the DV-based
representation is geared towards SAV.} However, AV and AA can (as
discussed below) be recast in terms of SAV, and vice-versa; as a
result, we will consider the two representations as general-purpose
alternatives, and we will study them as such.


\subsection{\diffvectors\ result in more training examples for AV}
\label{sec:moretrainingdata}

\noindent Our working hypothesis is that the DV-based representation
is advantageous. In order to show this, let us consider AV, and let us
assume that $A^{*}\in\mathcal{A}$ is our candidate author.
When using the standard representation, we typically replace each
label in $\mathcal{A}\setminus \{A^{*}\}$ with label
$\overline{A^{*}}$ (to indicate the complement of $A^{*}$) and train a
binary classifier that discriminates between $A^{*}$ and
$\overline{A^{*}}$. However, in doing so a lot of information is lost,
namely, the information whether two training examples in
$\overline{A^{*}}$ are by the same author or not. For
authorship-related tasks this is valuable information, which the
standard representation wastes and the DV-based representation does
not. The following example shows that the information wasted by the
standard representation is, indeed, \emph{a lot}.

\begin{example}
  \label{ex:one}
  Assume a set of 10 authors and a training set consisting of 100
  training examples for each author. The DV-based representation gives
  rise to (1,000$\cdot$999)/2=499,500 \dv, among which:
  \begin{enumerate}

  \item \label{item:same} 10$\cdot$(100$\cdot$99)/2=49,500 examples
    have label \textsf{Same}, since for each author
    $A_{z}\in\{A_{1},\ldots,A_{10}\}$ there are (100$\cdot$99)/2=4,950
    unordered pairs of different examples such that the author of both
    examples is $A_{z}$; of these
 
    \begin{enumerate}

    \item \label{item:same1} (100$\cdot$99)/2=4,950 are such that the
      author of both examples is $A^{*}$;

    \item \label{item:same2} 9$\cdot$(100$\cdot$99)/2=44,550 are such
      that the author of both examples is $A_{z}$ for some
      $A_{z}\not = A^{*}$;

    \end{enumerate}

  \item \label{item:diff} 45$\cdot$(100$\cdot$100)=450,000 examples
    have label \textsf{Different}, since there are 10$\cdot$9/2=45
    unordered pairs $(A',A'')$ of different authors, and for each such
    pair there are 100$\cdot$100=10,000 pairs of examples in which one
    example is by $A'$ and the other example is by $A''$; of these
 
    \begin{enumerate}

    \item \label{item:diff1} 9$\cdot$(100$\cdot$100)=90,000 are such
      that one of $A'$ and $A''$ is $A^{*}$;

    \item \label{item:diff2} 36$\cdot$(100$\cdot$100)=360,000 are such
      that neither of $A'$ and $A''$ is $A^{*}$.

    \end{enumerate}

  \end{enumerate}

  \noindent Note that the information provided to the training process
  by the examples of Type~\ref{item:same1} is also provided (albeit in
  a different form) when using the standard representation, since with
  the latter the learner is implicitly told that the two documents are
  from the same author. The same happens for the examples of
  Type~\ref{item:diff1}, since with the standard representation the
  learner is implicitly told that the two documents are from different
  authors. However, the key observation here is that the examples of
  Type~\ref{item:same2} and Type~\ref{item:diff2} provide information
  that is instead lost when using the standard representation, since
  the standard representation only tells the learner that the two
  documents are not by $A^{*}$, but does not tell the learner if they
  are by the same author or not. In sum, 404,550 out of 499,500
  training examples, i.e., about 81\% of the entire set, provide
  information that was not provided by the standard representation; in
  other words, in this case the learner receives more than 5 times the
  amount of information than the standard representation provides to
  it. \qed
 
\end{example}
 
More in general, if we have $m$ authors and $q=n/m$ training examples
per author, the number of DVs that do \emph{not} provide additional
information with respect to the standard representation is
\begin{align}
  \begin{split}
    \label{eq:noadditional}
    \frac{q(q-1)}{2}+(m-1)q^{2}
  \end{split}
\end{align}
\noindent i.e., the number of pairs of Type \ref{item:same1} plus the
number of pairs of Type \ref{item:diff1},
while the number of DVs that do provide additional information is
\begin{align}
  \begin{split}
    \label{eq:additional}
    \frac{(m-1)q(q-1)}{2}+\frac{(m-1)(m-2)q^{2}}{2}
  \end{split}
\end{align}
\noindent i.e., the number of pairs of Type \ref{item:same2} plus the
number of pairs of Type \ref{item:diff2}. Note that, while the amount
of information that was already available to the learning process is
$O(mq^2)$ (Equation~\ref{eq:noadditional}), the new information made
available to it is $O(m^2 q^2)$ (Equation~\ref{eq:additional}). The
latter amount of information can be extremely valuable, especially
since it comes at no cost, and especially in application scenarios (as
there are many in authorship identification) characterised by the
scarcity of training data. Among all of the above,
\begin{align}
  \begin{split}
    \frac{q(q-1)}{2}+\frac{(m-1)q(q-1)}{2}=\frac{mq(q-1)}{2}
  \end{split}
\end{align}
\noindent are examples of \textsf{Same}, which are $O(mq^{2})$, while
\begin{align}
  \begin{split}
    (m-1)q^{2}+\frac{(m-1)(m-2)q^{2}}{2} 
    = \frac{m(m-1)q^2}{2}
  \end{split}
\end{align}
\noindent 
are examples of \textsf{Different}, which are $O(m^{2}q^{2})$.

In sum, when our task is AV, if we switch from standard
representations to DV-based representations, we end up with a much
higher quantity of training data, since DV-based representations
exploit information that standard representations waste. Hovewer, note
that switching from standard representations to DV-based
representations means switching (as noted at the end of
Section~\ref{sec:diff}) from vectors geared towards AV to vectors
geared towards SAV. This suggests the idea to use these vectors to
train a high-performance SAV classifier, and then to devise an
algorithm that can perform AV on top of this SAV classifier; this is
the goal we will pursue in Section~\ref{sec:methodav}.

\subsection{\diffvectors\ make training more robust in closed-set AA}
\label{sec:morerobusttraining}

\noindent The fact that more information is provided to the training
process holds for AV, but does not necessarily hold for other
authorship identification tasks. In general, this fact only holds for
tasks in which, as in AV, the training documents by different authors
end up being grouped together into a single class; this happened in AV
with the $\overline{A^{*}}$ class.
%
However, that more information is provided to the training process
does not hold when the above-mentioned grouping does not happen, as,
e.g., in closed-set AA. In the latter task, the information conveyed
to the training process by a DV with label \textsf{Same} is obviously
also implicitly conveyed when using the standard representation (where
the vectors corresponding to the two documents are labelled with the
same author), and the same holds for DVs with label \textsf{Different}
(where the two vectors are labelled with different authors).

So, in closed-set AA (and in the latter tasks in general) it would
appear that there is no advantage in using DVs. This is actually not
true, because the advantage is in the fact that, \emph{when using \dv,
all the training information is concentrated on labelling just two
classes}, i.e., \textsf{Same} and \textsf{Different}, while in the
classical representation this information is spread out thin, i.e., it
is used for labelling $m$ different classes, each of which thus ends
up having a smaller number of positive training examples. The
following example makes the point more concrete.
\begin{example}
  \label{ex:two}
  Assume we are dealing with closed-set AA; assume a set of $m=10$
  authors and a training set consisting of $q=20$ training examples
  for each author. The standard representation gives rise to
  $q \cdot m=200$ training vectors, 20 for each class, while the
  DV-based representation gives rise to $mq(mq-1)/2$=19,900 training
  vectors, among which 
  10$\cdot$(20$\cdot$19)/2=1,900 DVs for class \textsf{Same} and
  $10 \cdot 9 \cdot 20^2 / 2$ = 18,000 DVs for class
  \textsf{Different}. \qed
%
\end{example}

More in general, if we have $m$ authors and $q$ training examples per
author, in closed-set AA we have $mq(q-1)/2$ DVs of class
\textsf{Same} and $m(m-1)q^{2}/2$ DVs of class \textsf{Different},
which means that the ratio between the number of training examples of
\textsf{Same} and the number of training examples of
\textsf{Different} is
$$\displaystyle\frac{q-1}{q(m-1)}\approx \frac{1}{m-1}$$
This indicates that we are in the presence of an \emph{imbalanced}
binary classification problem (which is even more imbalanced if $m$ is
large); however, this is not a problem because, since we typically
have many training DVs (see e.g., Example~\ref{ex:two}), we can
subsample class \textsf{Different}, i.e., remove some among its many
training examples from the training set.
 

In sum, the use of the DV-based representation in closed-set AA allows
the SAV binary classifier to be trained robustly, thanks to the fact
that the existing amount of training information can be devoted to
solving a comparatively easier binary classification task rather than
a comparatively more difficult 1-of-$m$ classification task. We can
thus expect to obtain accurate SAV classification predictions; in
Section~\ref{sec:methodaa} we will see that these SAV predictions can
also be used by a downstream process to solve authorship
identification tasks such as AV and AA.
%

%
%


\section{Solving SAV, AA, and AV, by means of \diffvectors}
\label{sec:recasting}

%
%

\noindent One difference between the standard representation, in which
class labels represent authors, and the representation based on \dv,
in which class labels are in \{\textsf{Same},\textsf{Different}\}, is
that the tasks that can be solved ``directly'' are AV and AA for the
former, and SAV for the latter. That is, by using the standard
representation, AV and AA can be solved directly by setting up a
classifier that, for a given document, returns a class label in
$\mathcal{A}$ (for AA) or in $\{A^{*},\overline{A}^{*}\}$ (for AV);
SAV is instead to be solved as a derivative, ``downstream'' task,
e.g., by first determining the true authors of documents $x_{i}$ and
$x_{j}$ by means of two calls to an AA engine, and then checking
whether the two returned class labels are the same or
not.\footnote{This is possible only for closed-set SAV, though, since
open-set SAV cannot be recast in terms of AA.
} On the contrary, when using the DV-based representation, SAV is
solved directly; AV and AA are instead to be solved as derivative
tasks, using SAV as the building block of any algorithm for solving
them. In this section we first formally define our method for
performing SAV (Section \ref{sec:methodsav}), and then go on to
describe two alternative solutions for solving both AV and AA (Section
\ref{sec:methodaa}) that build on top of the former.


\subsection{Solving SAV by means of \diffvectors}
\label{sec:methodsav}

\noindent Given a training set
$\mathcal{L}=\{(x_{1},y_{1}), ..., (x_{n},y_{n})\}$ of documents
$x_{i}\in\mathcal{D}$ labelled by classes
$y_{i}\in\mathcal{A}=\{A_{1}, ..., A_{m}\}$ representing authors, we
define its \emph{pair-based version} as
\begin{align}
  \begin{split}
    \label{eq:lprime}
    \mathcal{L}_{\mathcal{P}}=\{((x_{i},x_{j}),\mathrm{SD}(y_{i},y_{j}))
    \ | \ i,j\in\{1,..., n\}, j< i\}
  \end{split}
\end{align}
%
\noindent where $\mathrm{SD}(y_{i},y_{j})$
is an indicator function that returns \textsf{Same} if $y_{i}=y_{j}$
and \textsf{Different} otherwise.
We also assume a feature extractor
$f:\mathcal{D}\rightarrow \mathbb{R}^{t}$ which maps documents
$x\in\mathcal{D}$ into $t$-dimensional vectors $\mathbf{x}$ of real
numbers. We can thus rewrite $\mathcal{L}$ as
$\{(\mathbf{x}_{1},y_{1}), ..., (\mathbf{x}_{n},y_{n})\}$ and redefine
$\mathcal{L}_{\mathcal{P}}$ as
\begin{align}
  \begin{split}
    \label{eq:lprime2}
    \mathcal{L}_{\mathcal{P}}=\{(\mathbf{x}_{ij},\mathrm{SD}(y_{i},y_{j}))
    \ | \ i,j\in\{1,..., n\}, j< i\}
  \end{split}
\end{align}
\noindent where $\mathbf{x}_{ij}\in\mathbb{R}^{t}$ is a vector of
absolute differences of feature values, i.e., $\mathbf{x}_{ij}$ is the
vector such that its $k$-th component is
$\mathbf{x}_{ij}^{k}=|\mathbf{x}_{i}^{k}-\mathbf{x}_{j}^{k}|$, for all
$1\leq k \leq t$, $i\neq j$.

Note that $|\mathcal{L}|=n$ while
$|\mathcal{L}_{\mathcal{P}}|=n(n-1)/2$, i.e., the pair-based version
$\mathcal{L}_{\mathcal{P}}$ is $(n-1)/2$ times larger than its
standard counterpart $\mathcal{L}$. In practice, the size of
$\mathcal{L}_{\mathcal{P}}$ can be so large as to make the learning
process intractable for some batch learners.
For example, the 499,500 training DVs of Example~\ref{ex:one} would
result from a dataset of 10 authors and 100 training documents per
author, which is not a terribly large dataset. As shown in
Section~\ref{sec:morerobusttraining}, $\mathcal{L}_{\mathcal{P}}$
tends to be imbalanced,
with a \textsf{Same} / \textsf{Different} training example ratio
close, assuming a training set containing the same number of documents
for each author, to $1/m$.


In practice, we will be interested in generating and using only a
subset $\mathcal{L}'_{\mathcal{P}}\subset \mathcal{L}_{\mathcal{P}}$:
by including in $\mathcal{L}'_{\mathcal{P}}$ a small enough number of
elements of $\mathcal{L}_{\mathcal{P}}$ we can make the training
process tractable, and by including in $\mathcal{L}'_{\mathcal{P}}$ an
equal number of examples of \textsf{Same} and \textsf{Different} we
can avoid the typical negative consequences of imbalance.
By using a subset $\mathcal{L}'_{\mathcal{P}}$ with these
characteristics, we can then train a binary classifier
$h: \mathbb{R}^{t} \rightarrow
\{\textsf{Same},\textsf{Different}\}$. We can this classifier DV-Bin,
since it is a \underline{bin}ary classifier that uses DVs.

%

Without loss of generality, and for ease of notation, we will
henceforth use $h$ as the function of two arguments
%
$h:\mathcal{D}\times\mathcal{D}\rightarrow\{\textsf{Same}$,
\textsf{Different}\},
%
thus leaving implicit the phases of (a) mapping documents to feature
vectors, and (b) computing DVs from the absolute differences of
feature values. As a result, we can simply write $h(x_{i},x_{j})$ to
indicate a predicted label in $\{\textsf{Same},\textsf{Different}\}$.



\subsection{Solving AA by means of \diffvectors}
\label{sec:methodaa}


\noindent In this section we describe how SAV can be used to implement
AA as downstream tasks.


In order to predict by whom among the authors in $\mathcal{A}$ a test
document $x$ has been written, and to do so by using a SAV classifier,
it makes sense to look at how $x$ relates to the training documents in
terms of the \textsf{Same} and \textsf{Different} classes. For
instance, if for all documents $x'\in \mathcal{L}$
written by $A_{z}$ the pair $(x,x')$ is assigned by the SAV classifier
to class \textsf{Same}, and if for all $x''\in \mathcal{L}$ written by
an author in $\mathcal{A}\setminus \{A_{z}\}$ the pair $(x,x'')$ is
assigned to class \textsf{Different}, it would be reasonable to
predict that $x$ has been written by $A_{z}$.

Unfortunately, this uniformity rarely occurs in practice: in more
typical cases the SAV classifier will assign to class \textsf{Same},
say, some pairs $(x,x')$ where $x'$ has been written by $A_{z}$, and
some pairs $(x,x'')$ where $x''$ has been written by an author other
than $A_{z}$.
%
This brings up the question: how should we act in the presence of such
apparently contradictory outcomes?

Given that we need to build our AA algorithm on top of the output of
the SAV classifier, it is in our best interest to squeeze every
possible bit of information from this output. As a result, we will be
interested in exploiting not just the binary prediction of the SAV
classifier, but also its non-binary classification score, representing
the degree of certainty with which it has issued this prediction. We
assume that our SAV classifier is of the form
\begin{align}
  \begin{split}
    \label{eq:classifierhsoft}
    h:\mathcal{D}\times\mathcal{D}\rightarrow [0,1]
  \end{split}
\end{align}
\noindent i.e., returns classification scores that are \emph{posterior
probabilities}. These latter are values
$\Pr(\textsf{Same}|x_{i},x_{j})$ that denote the probability that the
SAV classifier attributes to the fact that $x_{i}$ and $x_{j}$ have
been written by the same author, and are such that
$\Pr(\textsf{Different}|x_{i},x_{j})=1-\Pr(\textsf{Same}|x_{i},x_{j})$.

We explore two techniques for building AA classifiers on top of SAV
classifiers, one inspired by \emph{lazy learning} methods
\cite{Aggarwal:2014hb} and another inspired by the well known
\emph{Stacked Generalisation} algorithm \cite{Wolpert:1992rq}.


\subsubsection{Lazy AA}
\label{sec:knn}

\noindent The first SAV-based AA algorithm that we explore in this
paper, and that we call \emph{Lazy AA}, draws inspiration from
distance-weighted $k$-NN, but is different from it. Similarly to
distance-weighted $k$-NN, the underlying idea of our method is that,
given a test document $x$, if a training document $x'$ authored by
$A_{z}$ is ``stylistically similar'' to $x$, this brings evidence
towards the fact that also $x$ is authored by $A_{z}$, and this
evidence can be quantified exactly by the amount of stylistic
similarity.
Differently from distance-weighted $k$-NN, though, instead of having
access to a function that computes the similarity between two
documents, we here have access to a SAV (soft) classifier that
computes the probability that the two documents are in class
\textsf{Same}. It is thus just natural to compute the stylistic
similarity between $x$ and $x'$ as $\Pr(\textsf{Same}|x,x')$, i.e., as
the probability that the SAV classifier attributes to the fact that
$x$ and $x'$ have been written by the same author.

Our combination rule thus consists of selecting, for each author
$A_{z}\in\mathcal{A}$, the $k$ training documents written by $A_{z}$
that are stylistically most similar to our test document $x$ (i.e.,
the ones for which $\Pr(\textsf{Same}|x,x')$ is highest), and
computing the average value of this stylistic similarity across these
$k$ documents; the author for which this average stylistic similarity
is highest is predicted to be the author of $x$.
In symbols, this comes down to
\begin{align}
  \label{eq:dvknn}
  \begin{split}
    h'(x,\mathcal{L},k)= & \ \mathop{\arg\max}_{A_{z}\in\mathcal{A}}
    \frac{1}{k}\sum_{x_{i}\in \mathrm{NN}(k,\mathcal{L},A_{z},x,h)} h(x,x_{i}) \\
    = & \ \mathop{\arg\max}_{A_{z}\in\mathcal{A}}
    \frac{1}{k}\sum_{x_{i}\in \mathrm{NN}(k,\mathcal{L},A_{z},x,h)}
    \Pr(\textsf{Same}|x,x_{i})
  \end{split}
\end{align}
\noindent where $\mathrm{NN}(k,\mathcal{L},A_{z},x,h)$ returns the $k$
documents from training set $\mathcal{L}$ that have been written by
author $A_{z}$ and are closest to $x$ according to the SAV classifier
$h$.
Note that the $h'$ functional is parameterised by $\mathcal{L}$ (and
$k$) since, as in all lazy learning methods, there is no proper
training phase for $h'$,
and all the computation is carried out at classification time.

The optimal value for parameter $k$ can be found via ``leave-one-out''
(LOO) validation on the training set $\mathcal{L}$. That is, for each
value of $k$ in the tested range each training document
$x_{i}\in \mathcal{L}$ is classified by a classifier $h'$ trained on
$\mathcal{L}\setminus\{x_{i}\}$; $k$ is thus set to the value that
maximises a given effectiveness measure as computed on the entire set
$\mathcal{L}$.\footnote{One might wonder why we go for LOO, a
traditionally expensive (and sometimes \textit{too} expensive) way of
optimising parameters, rather than the cheaper $t$-fold
cross-validation ($t$-FCV). The reason is that, in our case, LOO is no
more expensive than $t$-FCV because we are in a \textit{lazy} learning
context. In other words, in traditional \textit{eager} learning
contexts LOO requires $|\mathcal{L}|$ classifier retrainings, while
$t$-FCV requires only $t\ll |\mathcal{L}|$ classifier retrainings;
however, in lazy learning contexts there are no retrainings because
classifiers are not ``trained'', since all inductive inference is
carried out at classification time.}
If we use \emph{(vanilla) accuracy} (i.e., the proportion of correctly
classified instances) as the effectiveness measure, this process comes
down to computing
\begin{equation}
  k^*=\mathop{\arg\max}_{k} \frac{1}{n}\sum_{(x_{i},y_{i})\in \mathcal{L}} 
  \mathbf{1}[h'(x_{i},\mathcal{L} \setminus \{x_{i}\},k)=y_{i}]
  \label{eq:koptim}
\end{equation}
\noindent where $\mathbf{1}[s]$ is an indicator function returning 1
if statement $s$ is true and 0 otherwise. This optimisation can be
performed very quickly if the posterior probabilities
$\Pr(\textsf{Same}|x_{i},x_{j})$ are computed only once for all
$x_{i},x_{j}\in \mathcal{L}$ and stored for fast reuse.
Similarly, $\mathrm{NN}(k,\mathcal{L},A_{z},x,h)$ can be made to
return the top $k$ elements (for different values of $k$) from a fully
ranked list that is computed once and reused when necessary. Note also
that the majority of these operations are amenable to parallelisation.






\subsubsection{Stacked AA}
\label{sec:stacking}

\noindent \emph{Stacked AA} (so called since it is inspired by
\textit{stacked generalisation} -- \citep{Wolpert:1992rq}) consists of
an AA (single-label multiclass) classifier $h'$, trained by
general-purpose learning algorithms, that classifies documents
represented by vectors of posterior probabilities
$\Pr(\textsf{Same}|x,x_{k})$, each of which has been returned by an
underlying, previously trained SAV classifier $h$ (more precisely, a
DV-Bin classifier of the type described in
Section~\ref{sec:methodsav}). More in detail, in order to predict who
among the authors in $\mathcal{A}$ has written document $x$, we
represent $x$ via a vector
\begin{align}
  \begin{split}
    \label{eq:stackedaa}
    \phi(x)= & \ (h(x,x_1), \ldots, h(x,x_n)) \\
    = & \ (\Pr(\textsf{Same}|x,x_{1}), \ldots,
    \Pr(\textsf{Same}|x,x_{n}))
  \end{split}
\end{align}
\noindent of $n$ posterior probabilities, one for each training
example in $\mathcal{L}$. The $k$-th value in this vector is the value
$h(x,x_{k})=\Pr(\textsf{Same}|x,x_{k})$, where $x_{k}$ is the $k$-th
training example. In other words, in order to classify $x$ we first
need to perform $|\mathcal{L}|$ SAV classifications, where the $k$-th
such classification attempts to predict whether the test document $x$
was written by the same author who also wrote training document
$x_{k}$.

At training time, we train the AA classifier $h'$ by using all the
training examples in $\mathcal{L}$ represented in the style of
Equation~\ref{eq:stackedaa}. In other words, by applying the mapping
$\phi:\mathbb{R}^{t}\rightarrow[0,1]^{n}$ to the training documents
themselves we define a new ``view''
$\mathcal{L}_{h}=\{(\phi(x_{i}),y_{i})\}_{i=1}^{n}$ of the training
set $\mathcal{L}$, in which the training documents are not represented
via vectors of $t$ stylometric features but, thanks to the underlying
SAV classifier, via vectors of $|\mathcal{L}|$ posterior
probabilities, with $\phi(x)\in[0,1]^{n}$. 
The training set $\mathcal{L}_{h}$ can directly be used to train a
general-purpose classifier $h':[0,1]^{n}\rightarrow\mathcal{A}$ in the
feature space of posterior probabilities.\footnote{Note that, if the
learning algorithm is a linear model, then it takes the form of
$h'(x)=\sum_{l=1}^n\alpha_l h(x,x_l)$, in which $\{\alpha_l\}_{l=1}^n$
are the parameters to be learned, and the set of functions
$\{h(\cdot, x_l)\}_{l=1}^n$ plays the role of a set of basis functions
centred at the training
points.} 
Of course, in order to generate
$\mathcal{L}_{h}=\{(\phi(x_{i}),y_{i})\}_{i=1}^{n}$ we first need to
train a SAV classifier $h$ via the DV-Bin method of
Section~\ref{sec:methodsav}. In the experiments of Section
\ref{sec:experiments} we will concentrate on instantiations of $h'$
that are generated by the same learning method (e.g., logistic
regression) used to generate $h$.

At classification time, a given test document $x$ is classified by
first computing $\phi(x)$ (this requires invoking $n$ times classifier
$h$) and then invoking classifier $h'(\phi(x))$.

There are several important aspects in which Stacked AA differs from
Lazy AA:
\begin{itemize}

\item in Stacked AA, evidence is provided by \textit{all} training
  examples, and not just by the $k$ examples most similar to the test
  example, as is instead the case in Lazy AA;

\item in Stacked AA, the combination rule (i.e., the rule that
  assembles the evidence provided by the training examples into a
  final decision) is \textit{learnt} by a metaclassifier, i.e., it is
  not static, as is instead the case in Lazy AA;

\item in Stacked AA, learning is performed offline (since the
  metaclassifier is trained before the testing phase begins), while in
  Lazy AA all inductive inference is carried out at classification
  time.

\end{itemize}

\noindent One important aspect in which Stacked AA differs from
Stacked Generalisation, instead, is that
in Stacked Generalisation the metaclassifier and the base classifiers
are \emph{homogeneous}, i.e., all use the same set of classes, while
in Stacked AA the metaclassifier and the base classifiers are
\emph{heterogeneous}, i.e., use different sets of classes. Indeed, the
base classifiers use the classes in
\{\textsf{Same},\textsf{Different}\}, since they are binary SAV
classifiers, while the metaclassifier use the classes in
$\mathcal{A}=\{A_{1}, ..., A_{n}\}$, since it is a single-label
multiclass AA classifier.
%


\subsection{Solving AV by means of \diffvectors}
\label{sec:methodav}

\noindent It is fairly straightforward to take the algorithms
described in Sections~\ref{sec:knn} and \ref{sec:stacking} and
generate versions (that we will dub \emph{Lazy AV} and \emph{Stacked
AV}) that solve AV instead of AA. The only difference between Lazy AV
and Lazy AA, and between Stacked AV and Stacked AA, is that in the AV
versions of the two algorithms the codeframe used is binary, i.e., it
is $\mathcal{A}=\{A^{*},\overline{A^{*}}\}$; in particular, this means
that for Stacked AV the metaclassifier $h'$ is a binary classifier
instead of a multiclass classifier. Everything else is unmodified.

However, in preliminary experiments that we have run, both Lazy AV and
Stacked AV proved substantially inferior to versions of Lazy AA and
Stacked AA, respectively, in which we attribute document $x$ to
$A^{*}$ if the AA algorithm does so and we attribute $x$ to
$\overline{A^{*}}$ if the AA algorithm attributes it to an author
$A_{z}$ different from $A$. Concerning the reason why Lazy AV
underperforms Lazy AA, this has likely to do with the fact that there
is an \textit{a priori} high probability that the $k$ nearest
neighbours in $\overline{A^{*}}$ are, on average, closer to $x$ than
the $k$ nearest neighbours in $A^{*}$, since $\overline{A^{*}}$ is a
very large pool to choose from (this does not happen in AA, where,
assuming an equal number of training documents per author, all pools
are equally large); this can give undue advantage to
$\overline{A^{*}}$ over $A^{*}$, and thus generate a large quantity of
false negatives. Concerning the reason why Stacked AV underperforms
Stacked AA, this has likely to do with the fact that the
metaclassifier of Stacked AV does not put the available class
information to the best use, i.e., conflates all labels different from
$A^{*}$ into a single label $\overline{A^{*}}$ that ends up being
poorly characterised from the semantic point of view.

Therefore, in the rest of the paper the algorithms we will use for
solving AV via DV-based representations will be the versions of Lazy
AA and Stacked AA described at the beginning of the previous
paragraph.
A consequence of this is that any AA experiment that involves the use
of either Lazy AA and Stacked AA and a codeframe
$\mathcal{A}=\{A_{1}, \ldots, A_{m}\}$, is also \textit{de facto} a
set of $m$ different AV experiments. In other words, we will not need
to run separate AA and AV experiments, i.e., we will evaluate the AA
experiments that we describe in Section~\ref{sec:experiments} both in
terms of AA \emph{and} AV.


\section{Experiments}
\label{sec:experiments}

\noindent In order to test whether a representation based on DVs is
advantageous with respect to a representation based on standard
vectors, we compare these two different design choices in experiments
that we run on four publicly available datasets (among which one that
we here make available for the first time) and for all three
authorship analysis tasks (AA, AV, SAV). The code to reproduce our
experiments is available online at
\url{https://github.com/AlexMoreo/diff-vectors} .


\subsection{Datasets}
\label{sec:Datasets}

\noindent We run experiments on four datasets consisting of textual
documents annotated by author; our datasets are representative of
different textual genres, lengths, and styles, are publicly available,
and all consist of English texts. The four datasets are:

\begin{itemize}

\item \texttt{IMDB62}. This dataset\footnote{Available at:
  \url{https://umlt.infotech.monash.edu/?page_{i}d=266}} was created
  and made publicly available (along with an extended version,
  \texttt{IMDB1million}) by \citet{Seroussi:2014jn}. It contains film
  reviews collected from the popular Internet Movie Database, and
  accounts for 62 authors/reviewers and 1,000 reviews authored by each
  of them. In order to divide the 62,000 documents into a training set
  and a test set we perform a stratified split, resulting in 700
  training documents and 300 test documents for each author. We use
  these texts as examples of a ``moderately formal'' type of
  communication, since the reviews are not as short as, for example,
  online messages, and, despite some occasional slang, are written in
  a clear and correct (although often informal) manner.

\item \texttt{PAN2011}. This dataset\footnote{Available at:
  \url{https://pan.webis.de/clef11/pan11-web/authorship-attribution.html}}
  was created for the PAN 2011 international authorship identification
  competition \cite{Argamon:2011wu}. The dataset is based on the Enron
  email corpus \cite{Klimt:2004tv}, i.e., the documents are emails
  annotated by author. \citet{Klimt:2004tv} have removed personal
  names and email addresses and replaced them with specific tags,
  which means that an authorship identification method is not able to
  use this extremely revealing information. In our experiments we use
  the ``Large'' training set (containing 9,337 documents, altogether
  accounting for 72 different authors) and the corresponding test set
  (containing 1,300 documents altogether, by the same authors
  represented in the training set). The emails are often extremely
  short, and show many characteristics of online communication; in
  order to avoid texts which are excessively short (and thus too
  difficult to attribute), we remove emails consisting of fewer than
  15 words.

\item \texttt{Victorian}. This dataset\footnote{Available at:
  \url{https://archive.ics.uci.edu/ml/datasets/Victorian+Era+Authorship+Attribution}}
  was created and made publicly available by \citet{Gungor:2018tb}.
  It consists of books by American or English 18th-19th century
  novelists, subdivided into segments of 1,000 words each by the
  creators of the dataset. They also (i) removed the first and last
  500 words of each book, and, (ii) as a topic-filtering measure,
  retained only the occurrences of the 10,000 words most frequent in
  the dataset. The result is a corpus of more than 50,000 documents
  (i.e., segments) by 50 different authors; the corpus is an
  imbalanced one, with the least represented author accounting for 183
  segments and the most represented one accounting for about 4,000 of
  them. In order to divide it into a training set and a test set, we
  perform again a stratified split, including 70\% of each author's
  texts in the training set and the remaining 30\% in the test set. We
  use these documents as examples of literary production characterised
  by a sophisticated style.

\item \texttt{arXiv}. This dataset, which we have created and made
  publicly available ourselves,\footnote{Available at:
  \url{https://doi.org/10.5281/zenodo.7404702}} consists of abstracts
  of single-author papers from
  \texttt{arXiv}.\footnote{\url{https://arxiv.org/}} In order to limit
  domain-dependence we have harvested these abstracts by querying
  \texttt{arXiv}'s API with a list of computer-science-related
  keywords, mostly focused on machine learning.\footnote{The query
  used was ``deep learning, machine learning, information retrieval,
  computer science, data mining, support vector, logistic regression,
  artificial intelligence, supervised learning''.} Computer science
  articles are seldom written by a single author, which means that
  this dataset is not large. The corpus somehow follows a power-law
  distribution, with few prolific authors and many authors accounting
  for very few abstracts each: we retained authors with at least 10
  abstracts to their name, resulting in a total of 1,469 documents
  from 100 authors. The 2 most prolific authors have 34 abstracts to
  their name, the 10 most prolific authors have written 22 or more,
  while 50\% of the authors have no more than 12 abstracts to their
  name. In order to divide the corpus into a training set and a test
  set we performed a stratified split, with the production of each
  author being split into a training set (70\% of the abstracts) and a
  test set (30\%). We use these abstracts as examples of ``scientific
  communication", characterised by a precise and compact style, with
  an abundance of technical terminology.
\end{itemize}


\subsection{Learners}
\label{sec:Learners}

%
%
%
%
%
\noindent We use logistic regression (LR) as the learning method. LR
is a simple linear model that has delivered very good accuracy in a
number of text-related applications. LR has two further advantages,
i.e., (i) the classification scores returned by the classifiers
trained by it are posterior probabilities, and (ii) these
probabilities are \emph{well-calibrated}.\footnote{A well-calibrated
classifier is one that returns accurate posterior probabilities. An
intuition of what ``accurate posterior probabilities'' means can be
provided by the following example. If 10\% (resp., 90\%) of all the
documents $x_{i}$ for which $h(x_{i})=\Pr(A|x_{i})=0.5$ indeed belong
to class $A$, we can say that the classifier $h$ has overestimated
(resp., underestimated) the probability that these documents belong to
$A$, and that their posteriors are thus inaccurate. Conversely, if
this percentage is 50\%, we can say that the classifier $h$ has
correctly estimated the probability that these documents belong to
$A$, and that their posteriors are thus accurate. Indeed, we say (see
for instance \cite{Flach:2017zp}) that the posteriors
$h(x_{i})=\Pr(A|x_{i})$ are perfectly calibrated (i.e., accurate) with
respect to a (labelled) set $\sigma=\{(x_{i},y_{i})\}_{i=1}^{n}$ if,
for all $\alpha\in [0,1]$, it holds that
%
\begin{align}
  \label{eq:calibration}
  \frac{|\{(x_{i},y_{i})\in \sigma \mid
  h(x)=\alpha, y_{i}=A\}|}{|\{(x_{i},y_{i})\in
  \sigma \mid h(x)=\alpha\}|}=\alpha
\end{align}
\noindent The classifiers trained by means of some learners (and
logistic regression is one of them) are known to return reasonably
well-calibrated probabilities. Those trained by means of some other
learners (such as Naïve Bayes) return probabilities which are known to
be not well calibrated \cite{Domingos:1996fj}. Yet other learners
(such as SVMs or AdaBoost) train classifiers that return confidence
scores that are not probabilities (i.e., that do not range on [0,1]
and/or that do not sum up to 1). In order to address these two latter
cases, \emph{probability calibration} mechanisms exist (see e.g.,
\citep{Niculescu-Mizil:2005kx, Niculescu-Mizil:2005gh, Platt:2000fk,
Wu:2004rz, Zadrozny:2002eu}) that convert the outputs of these
classifiers into well calibrated probabilities.} These are important
advantages, since the methods we have described in
Sections~\ref{sec:knn} and~\ref{sec:stacking} do rely on posterior
probabilities, and obviously benefit from the fact that these
posteriors are high-quality.
%

We optimise the hyperparameter $C$ of LR (the inverse of the L2
regularisation strength) in the log-space $\{10^i\}_{i=0}^{i=4}$, and
select the value of $C$ that minimizes the multinomial loss in a
stratified $t$-fold cross-validation (with $t=5$).\footnote{We use the
\textsc{LogisticRegressionCV} \textsc{scikit-learn}'s implementation,
see
\url{https://scikit-learn.org/stable/modules/generated/sklearn.linear_model.LogisticRegressionCV.html}.}


In order to generate the \textsf{Same} and \textsf{Different} training
pairs, we adopt the following policy. Given a training set
$\mathcal{L}$, we first compute the number of \textsf{Same} pairs that
can be generated. If there are fewer than 50,000 \textsf{Same} pairs,
we generate them all; otherwise, we draw (uniformly at random) and
generate 50,000 \textsf{Same} pairs. We then draw (again, uniformly at
random) and generate as many \textsf{Different} pairs as the
\textsf{Same} pairs we have generated. This is in order to guarantee a
balanced training set, since there are usually many more potential
\textsf{Different} pairs than \textsf{Same} ones.


\subsection{Features}
\label{sec:features}

\noindent As for the choice of features, we stick to ones well-known
and broadly adopted in the field of authorship analysis, i.e.,
features of a frequentistic nature that can be extracted automatically
and that are believed to convey stylistic information; see for example
\cite{Eder:2011ju, Juola:2006fk, Stamatatos2009:yq} for an overview,
and \citet{Kestemont:2018or, Kestemont:2019ov} for a discussion of the
most frequently used features in recent shared tasks focused on
authorship analysis.

The features we use can be naturally subdivided into two groups. Group
1 is composed of
\begin{itemize}
\item \label{item:function} \emph{Function words.} Each function word
  that appears in the training set is a feature in our vectorial
  representations. We use the list of English function words provided
  by NLTK.\footnote{\url{https://www.nltk.org/}}
\item \emph{Word lengths.} Each word length instantiated in the
  training set is a feature.
%
\item \label{item:sentencel} \emph{Sentence lengths.} Each sentence
  length instantiated in the training set is a
  feature. 
\item \emph{Punctuation symbols.} Each punctuation symbol that occurs
  in the training set is a feature.
\end{itemize}
\noindent while Group 2 is composed of
\begin{itemize}
\item \label{item:POS} \emph{POS $n$-grams.} We extract parts of
  speech from our texts by using the Spacy
  library,\footnote{\url{https://spacy.io/}} and we consider each POS
  $n$-gram (for $n\in[3,4]$) that occurs in the training set as a
  potential feature (where ``potential'' means ``barring feature
  selection'' -- see below).
\item \emph{Word uni-grams.} We consider each word that occurs in the
  training set as a potential feature.
\item \label{item:character} \emph{Character $n$-grams.} We consider
  each character $n$-gram (for $n\in[2,5]$) that occurs in the
  training set as a potential feature.
\end{itemize}
\noindent The features in Group 1 (i) are relatively few (typically:
$O(10^{2})$), and (ii) are dense, i.e., all of them can be expected to
occur to some degree in most texts. Given one of these features and
given a document, as the value of the feature in the document we take
its relative frequency in the document; for instance, the value of
punctuation symbol ``!'' in a document will be the number of times
symbol ``!'' occurs in the document divided by the number of
punctuation symbols in the document. We also apply standardisation to
the columns that these features generate in the document-by-feature
matrix.\footnote{Standardisation (aka z-scoring) is a normalisation
process consisting of centring and scaling a random variable so as to
force its distribution to have 0-mean and 1-variance, i.e., the
z-score of a raw variable $x$ is defined as $z=\frac{x-\mu}{\sigma}$
where $\mu$ and $\sigma$ are the (sample) mean and (sample) standard
deviation of $x$ as estimated in the training set. For the benefits in
accuracy deriving from standardising dense features, see
\citep{Moreo:2018db}.}

The features in Group 2 are many (typically: $O(10^{4})$ or
$O(10^{5})$). In order to deal with the fact that they may be
\emph{too} many, we apply to them filter-style feature selection,
using the chi-square test as the term scoring function~\cite{Yang97}
and retaining the 50,000 highest-scoring features. As the feature
weighting function, rather than using plain relative frequency we use
tfidf (an increasing function of relative frequency) in its standard
``ltc'' variant (see e.g., \cite{Salton88}).\footnote{Using tfidf
(which is indeed an increasing function of relative frequency) for
weighting sparse features is customary in authorship analysis (see
e.g., \cite{Menta:2021by, Ikae:2021ne, Koppel:2014bq,
Koppel:2009ix}). This function is the combination of the tf factor,
which is somehow akin to relative frequency, with the idf factor,
which lends a higher weight to features that are rare in the training
set (see \cite{Salton88} for details); in authorship analysis, the use
of idf is justified by the fact that rare POS $n$-grams / word
unigrams / character $n$-grams can be considered more indicative of
style than common ones} The features in Group 2 are sparse, i.e., in a
given document a large number of them will not occur in it; we do not
apply any standardisation to the features in Group 2 since this would
turn them into dense features, and this would be detrimental to
efficiency.


\subsection{Intrinsic evaluation of \diffvectors}
\label{sec:intrinsic}

\noindent Our ``intrinsic'' evaluation of DVs consists of SAV
experiments, since SAV is the task that a classifier using DVs can
solve directly. In these experiments we use a set of authors
$\mathcal{A}=\{A_{1}, \ldots, A_{m}\}$, with $m>2$, each one being the
author of $q$ documents. Given a dataset that contains a test set
$\mathcal{U}$, we test our systems on randomly drawn samples of test
document pairs. The reason why we do not test on \emph{all} possible
pairs is (see also Section~\ref{sec:moretrainingdata}) a practical
one, i.e., the fact that the number $|\mathcal{U}|(|\mathcal{U}|-1)/2$
of all possible pairs is too high for all but the most trivial
datasets.
We randomly draw balanced subsets of 1,000 test pairs (500 positive
and 500 negative) for each experiment.

We investigate the impact on performance of the number $m$ of authors
and the number $q$ of training documents per author. Specifically, for
the \texttt{IMDB62}, \texttt{PAN2011}, and \texttt{Victorian} datasets
we run experiments varying the number $m$ of authors in the set
$\{5,10,15,20,25\}$ and the number $q$ of documents per author in the
set $\{10,20,30,40,50\}$. Samplings are incremental, i.e., we do not
resample from scratch; in other words, when moving from, say, $q=20$
to $q=30$, we add 10 new documents per author to the previous
20. Regarding the test set, for each choice of $m$ we draw 2,000
random test pairs, 1,000 of which consist of texts written by some
among the $m$ authors present in the training set (\emph{closed-set
SAV}), while the other 1,000 pairs consist of texts written by $m$
authors other than the $m$ authors present in the training set
(\emph{open-set SAV}).\footnote{As detailed in
Section~\ref{sec:authorshipanalysis}, in open-set SAV one normally
assumes that the authors of the two unlabelled documents are \emph{not
necessarily} among the authors represented in the training set; in
these experiments we consider the more difficult setting in which the
authors of the two unlabelled documents are \emph{strictly not} among
the authors represented in the training set.}

In order to compensate for the random effect introduced by sampling
(authors, documents, and test pairs), we report results obtained by
averaging across 10 runs for each combination (dataset, $m$, $q$); we
use the same random samples for all the methods we compare. The only
exception is the \texttt{arXiv} dataset, which, due to its limited
size, does not allow this extraction of multiple samples; hence, for
this dataset we simply report experiments across 10 random train/test
splits of the entire dataset.


We perform experiments in both closed-set SAV (Section
\ref{sec:sav:close}) and open-set SAV settings (Section
\ref{sec:sav:open}). We evaluate the performance in terms of vanilla
accuracy (fraction of correctly classified pairs), which is a
perfectly valid evaluation measure when the test set is balanced
across the classes, such as the present one.


\subsubsection{Experiments on closed-set SAV} \label{sec:sav:close}

\noindent In the closed-set scenario, the authors in the test set
$\mathcal{U}$ are the same as in the training set. We here explore two
variants of our method:

\begin{itemize}

\item DV-Bin: the binary classifier discussed in Section
  \ref{sec:methodsav}.
 
\item DV-2xAA: a method that solves SAV by building on top of the Lazy
  AA method
  discussed in Section \ref{sec:knn}. In other words, this method
  first predicts, for both unlabelled documents, who the author of the
  document is, and then checks if the two predicted authors are the
  same author.
\end{itemize}

\noindent We consider the following baseline systems:

\begin{itemize}

\item STD-CosDist: This consists of a binary classifier trained to
  predict whether the pair belongs to \textsf{Same} or
  \textsf{Different}, where a pair of documents is represented by a
  vector of one feature only. The value of this feature is obtained by
  calculating the distance between the two documents, each represented
  by a ``standard'' vector, and where the distance function is the
  cosine distance. We have also run experiments using the L1 or L2
  distances in place of the cosine distance; we omit to report their
  results since cosine proved the best-performing one. The training
  set is transformed into pairs following the same policy as in DV-Bin
  (see Section~\ref{sec:Learners}). The classifier thus learns the
  distance threshold that best separates the \textsf{Same} pairs from
  the \textsf{Different} pairs.
 
\item STD-2xAA: This consists of a single-label multiclass classifier
  that operates on standard vector representations and that, as in
  DV-2xAA, solves SAV by performing closed-set AA for both documents
  and then checking if the two predicted authors are the same.
 
\end{itemize}

\noindent Both baselines are equipped with the same learner as our
method, i.e., LR optimised by running the usual optimisation process
for hyperparameter $C$.

\begin{figure}[t]
  \begin{center}
    \includegraphics[width=.85\textwidth]{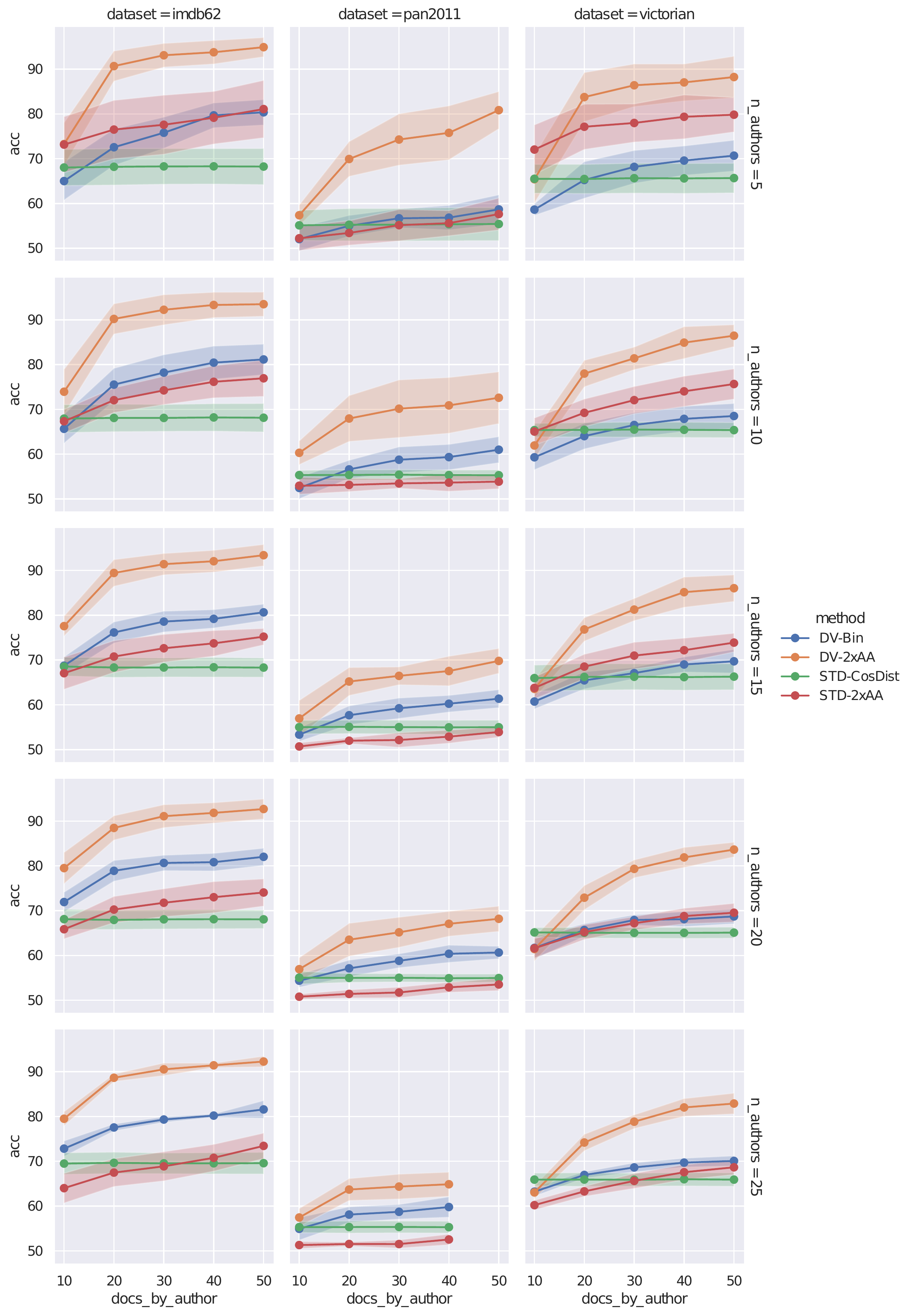}
 \caption{Intrinsic evaluation of \dv: results on closed-set
 SAV, using vanilla accuracy (on the $y$ axis) as the evaluation 
 measure on datasets \texttt{IMDB62}, \texttt{PAN2011}, and
\texttt{Victorian}. 
}
\label{fig:acc-close}
\end{center}
\end{figure}

Figure \ref{fig:acc-close} reports the experimental results we have
obtained, displayed in terms of accuracy (on the $y$ axis) as a
function of the number of training documents per author (on the $x$
axis), in datasets \texttt{IMDB62}, \texttt{PAN2011}, and
\texttt{Victorian} (each corresponding to a different column), at
varying number of authors (each corresponding to a different row). The
values for combination (\texttt{PAN2011},25,50) are missing since this
combination is not feasible, given that in \texttt{PAN2011} there are
fewer than 50 authors (25 for the closed-set setting and 25 for the
open-set setting) with at least 50 training documents each. Coloured
dots each represent an average result across 10 experiments, while the
colour band frontiers indicate $\pm$ one standard deviation from the
mean. Table \ref{tab:sav:arxiv:close} reports the results for the
\texttt{arXiv} dataset.

\begin{table}[t]
  \begin{tabular}{|l|rrl|}
    \hline
    & \multicolumn{1}{c}{mean} & \multicolumn{1}{c}{std} & 
                                                           \multicolumn{1}{c|}{ttest} \\
    \hline
    DV-Bin & .756 & 1.689 & \\
    DV-2xAA & \textbf{.803} & 2.545 & \\
    STD-CosDist & .629 & 2.241 & \\
    STD-2xAA & .646 & 1.430 & \\
    \hline
  \end{tabular}
  \caption{Intrinsic evaluation of \dv: results on closed-set SAV,
  using vanilla accuracy as the evaluation measure on dataset
  \texttt{arXiv}. \textbf{Boldface} indicates the best method. Symbols
  * and ** denote the method (if any) whose score is \emph{not}
  statistically significantly different from the best one at
  $\alpha=0.05$ (*) or at $\alpha=0.001$ (**) according to a paired
  sample, two-tailed t-test. No symbols * and ** appear in this
  particular table since all differences are statistically
  significant.}
  \label{tab:sav:arxiv:close}
\end{table}

The results clearly indicate that the DV-based variants perform well;
of the two methods that achieve SAV by running AA on both documents
(i.e., the DV-2xAA and STD-2xAA methods), the DV-based method is
always better or much better than the standard vector-based method,
and the same happens of the two non-AA-based methods. The
top-performing method is unquestionably DV-2xAA, which always
outperforms (often by a very large margin) all others, for all numbers
$m$ of authors and for all numbers $q$ of training examples per
author. As for the reason why DV-2xAA outperforms DV-Bin, we
conjecture that this may happen because the Lazy AA method uses only
evidence conveyed by few relevant training documents (the $k$
documents most similar to the test document, for both test documents),
thus filtering out other less relevant documents; this is in keeping
with the fact that methods based on nearest neighbours, as our DV-2xAA
method, always pick, during their parameter optimisation phase, values
of $k$ that are much smaller than the entire size of the training set.


All algorithms obviously improve their performance as the number of
documents per author increases, with the sole exception of
STD-CosDist. This latter fact might indicate that the optimal distance
threshold that STD-CosDist finds is fairly stable, and is well
estimated even by using few training data. However, it seems clear
from these results that distances alone do not carry as much
information as DVs do.




Figure \ref{fig:scores:close} shows the distribution of
$\Pr(\textsf{Same}|x',x'')$ values for \textsf{Same} and
\textsf{Different} pairs that STD-CosDist and DV-Bin compute. For this
experiment we have set $m=20$ and $q=50$ for all datasets except for
\texttt{arXiv}, where we have set $m=50$ and used all the documents
written by the 50 authors. (Note that the ``2xAA'' variants do not
compute a single posterior probability and are thus not amenable to a
similar analysis.) The STD-CosDist method manages to separate the
posteriors of the \textsf{Same} and \textsf{Different} pairs to some
extent in the \texttt{IMDB62} and \texttt{Victorian} datasets, but it
fails to separate them well in \texttt{PAN2011} and
\texttt{arXiv}. Interestingly enough, the posteriors generated by
STD-CosDist are close to being normally distributed, both for the
\textsf{Same} pairs and for the \textsf{Different} pairs.
Things are very different for the DV-Bin method, which tends to
generate much more polarised scores (i.e., separate the positives from
the negatives much better), placing most of the density mass around 0
for \textsf{Different} pairs and around 1 for \textsf{Same} pairs,
which is indicative of a very good performance. Still, the score
distribution generated for \texttt{Victorian} and, especially, for
\texttt{PAN2011}, reveal that the DV-Bin method still has room for
improvement.

\def \firstcolfile {Dist-Cos_close} \def \firstmethod {STD-CosDist}

\def \secondcolfile {DiffVectors_close} \def \secondmethod {DV-Bin}

\def \captionlabel {Distribution of $\Pr(\textsf{Same}|x',x'')$ values
for \textsf{Same} and \textsf{Different} pairs as computed by
STD-DistCos (first column) and DV-Bin (second column).} \def
\plotlabel{fig:scores:close} \def \datasetA {imdb62}
\def \datasetB {pan2011}
\def \datasetC {victorian}
\def \datasetD {arXiv}

\def \datasetnameA {imdb62}
\def \datasetnameB {pan2011}
\def \datasetnameC {victorian}
\def \datasetnameD {arXiv}

\begin{figure}
\centering

\begin{minipage}{.1\textwidth}
\center \scriptsize \side{IMDB62}
\end{minipage}
\begin{minipage}{.4\textwidth}
\center \scriptsize{\hspace{.5cm}\firstmethod} \includegraphics[trim={0 0 0 .9cm},clip,width=\linewidth]{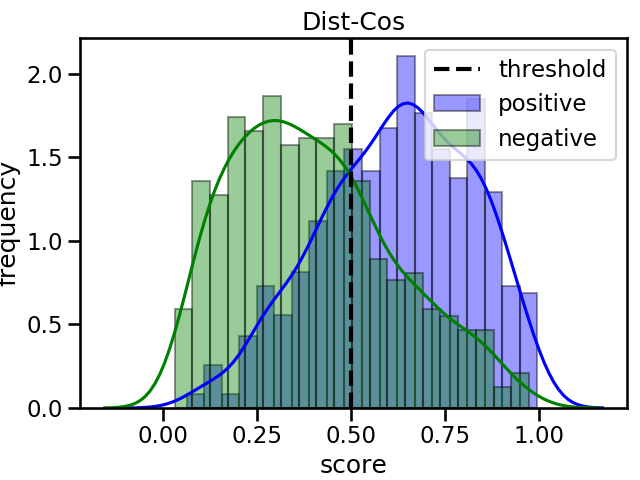}
\end{minipage}
\begin{minipage}{.4\textwidth}
\center \scriptsize{\hspace{.5cm}\secondmethod}
\includegraphics[trim={0 0 0 .9cm},clip,width=\linewidth]{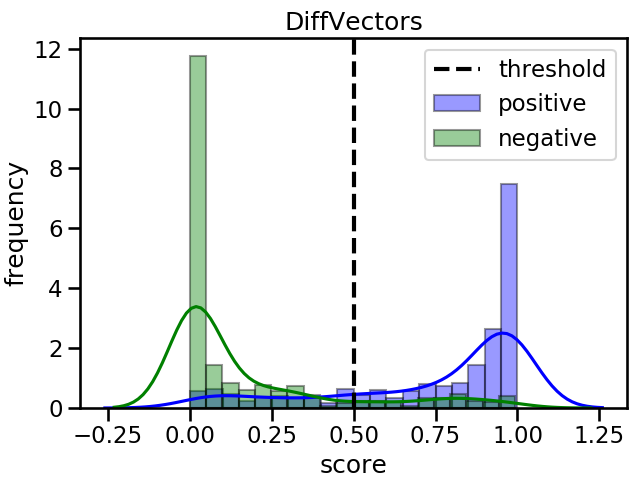}
\end{minipage}

\begin{minipage}{.1\textwidth}
\center \scriptsize \side{Pan2011}
\end{minipage}
\begin{minipage}{.4\textwidth}
\center \scriptsize{\hspace{.5cm}\firstmethod} \includegraphics[trim={0 0 0 .9cm},clip,width=\linewidth]{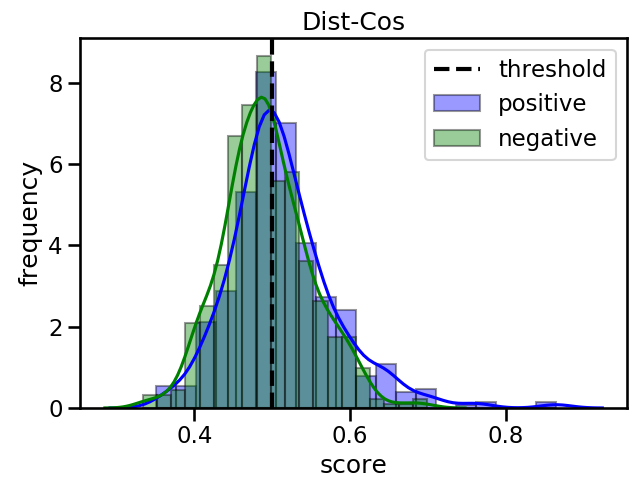}
\end{minipage}
\begin{minipage}{.4\textwidth}
\center \scriptsize{\hspace{.5cm}\secondmethod}
\includegraphics[trim={0 0 0 .9cm},clip,width=\linewidth]{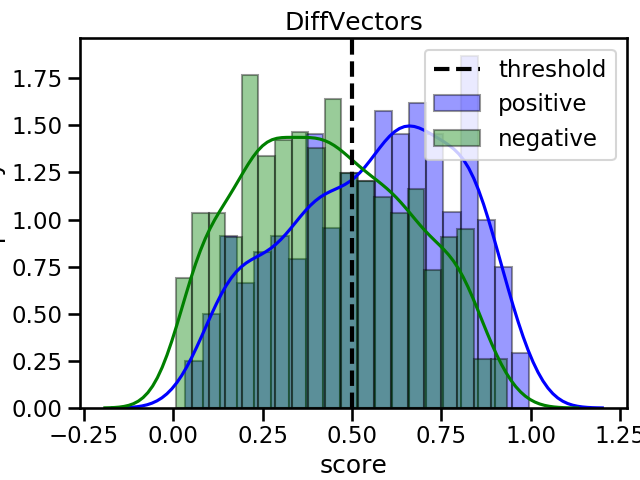}
\end{minipage}

\begin{minipage}{.1\textwidth}
\center \scriptsize \side{Victorian}
\end{minipage}
\begin{minipage}{.4\textwidth}
\center \scriptsize{\hspace{.5cm}\firstmethod} \includegraphics[trim={0 0 0 .9cm},clip,width=\linewidth]{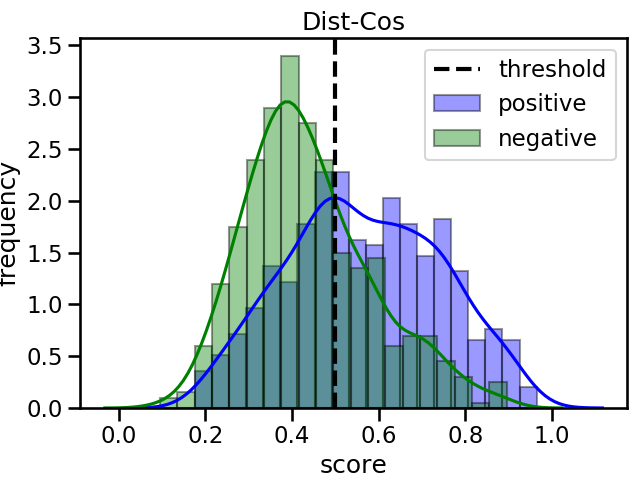}
\end{minipage}
\begin{minipage}{.4\textwidth}
\center \scriptsize{\hspace{.5cm}\secondmethod}
\includegraphics[trim={0 0 0 .9cm},clip,width=\linewidth]{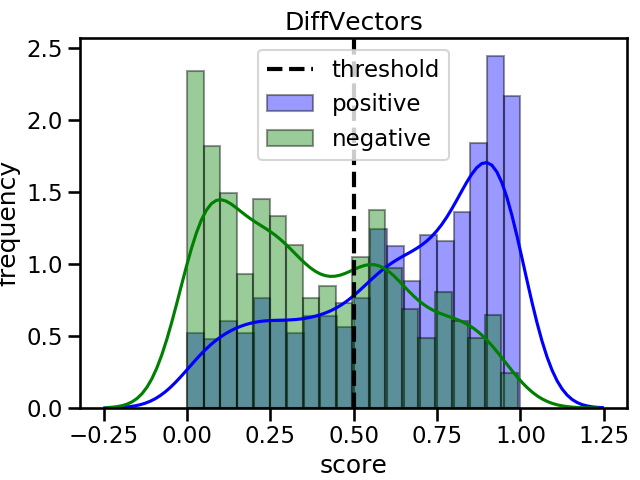}
\end{minipage}

\begin{minipage}{.1\textwidth}
\center \scriptsize \side{arXiv}
\end{minipage}
\begin{minipage}{.4\textwidth}
\center \scriptsize{\hspace{.5cm}\firstmethod} \includegraphics[trim={0 0 0 .9cm},clip,width=\linewidth]{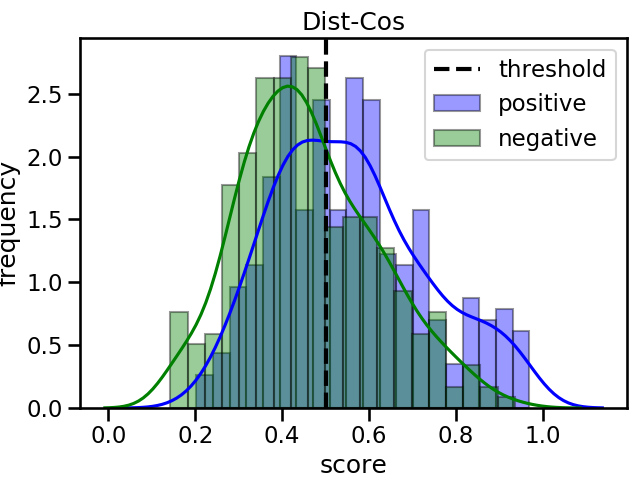}
\end{minipage}
\begin{minipage}{.4\textwidth}
\center \scriptsize{\hspace{.5cm}\secondmethod}
\includegraphics[trim={0 0 0 .9cm},clip,width=\linewidth]{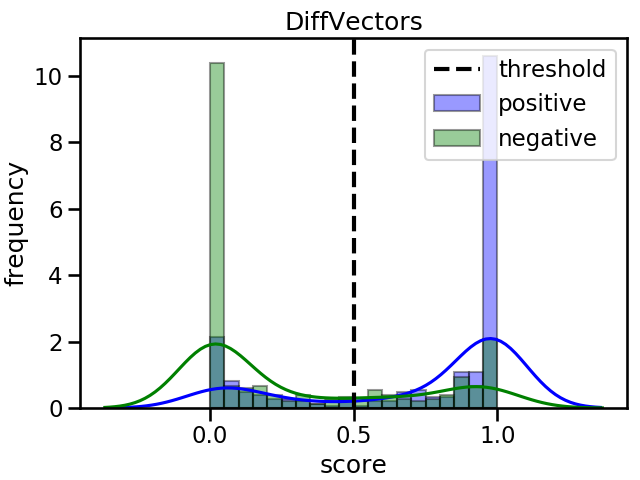}
\end{minipage}

\caption{\captionlabel}
\label{\plotlabel}
\end{figure}



\subsubsection{Experiments on open-set SAV}\label{sec:sav:open}

\noindent In the open-set SAV experiments, there is no intersection
between the set of $m$ authors that we draw to compose the test set
and the set of $m$ authors observed during training. This aspect
automatically rules out any attempt to perform SAV via authorship
attribution (i.e., DV-2xAA); for this reason, in this setting the only
DV-based method we test is DV-Bin. The baseline systems we consider
are:

\begin{itemize}

\item STD-CosDist: This is the same distance-based method that we have
  used in the closed-set SAV experiments. In this case the method is
  constrained to learn the optimal threshold from authors different
  from those in the test set.
 
\item Impostors: This is a method developed by \citet{Koppel:2014bq}.
  We use our own implementation of the ``blogger's'' variant, which
  had proved superior to others in the experiments of
  \cite{Koppel:2014bq} and amounts to using documents from the same
  domain (blogs in the original authors' experiments, documents from
  the training set in our case) as the impostors candidates. We use
  cosine as the distance function since in our experiments we have
  found it to consistently deliver better results than the ``minmax''
  criterion (the similarity function of choice in
  \cite{Koppel:2014bq}). We set parameter $I$ (the number of impostor
  candidates) to 50 instead of 250 (which was found to work well by
  \citet{Koppel:2014bq}) since our training sets are much smaller than
  those they considered (sticking to $m=250$ would basically result in
  a random choice of impostor candidates);
  following \cite{Koppel:2014bq}, the rest of the parameter values we
  use are $i=10$ (number of impostors) and $k=100$ (number of bagging
  trials). Also following \cite{Koppel:2014bq}, we optimise parameter
  $\sigma^*$ (the decision threshold) on a validation set. Note that
  we have not used this baseline method in the closed-set SAV
  experiments, since in that case the ``impostors'' cannot be created.
\end{itemize}







\begin{figure}[t]
  \begin{center}
    \includegraphics[width=.85\textwidth]{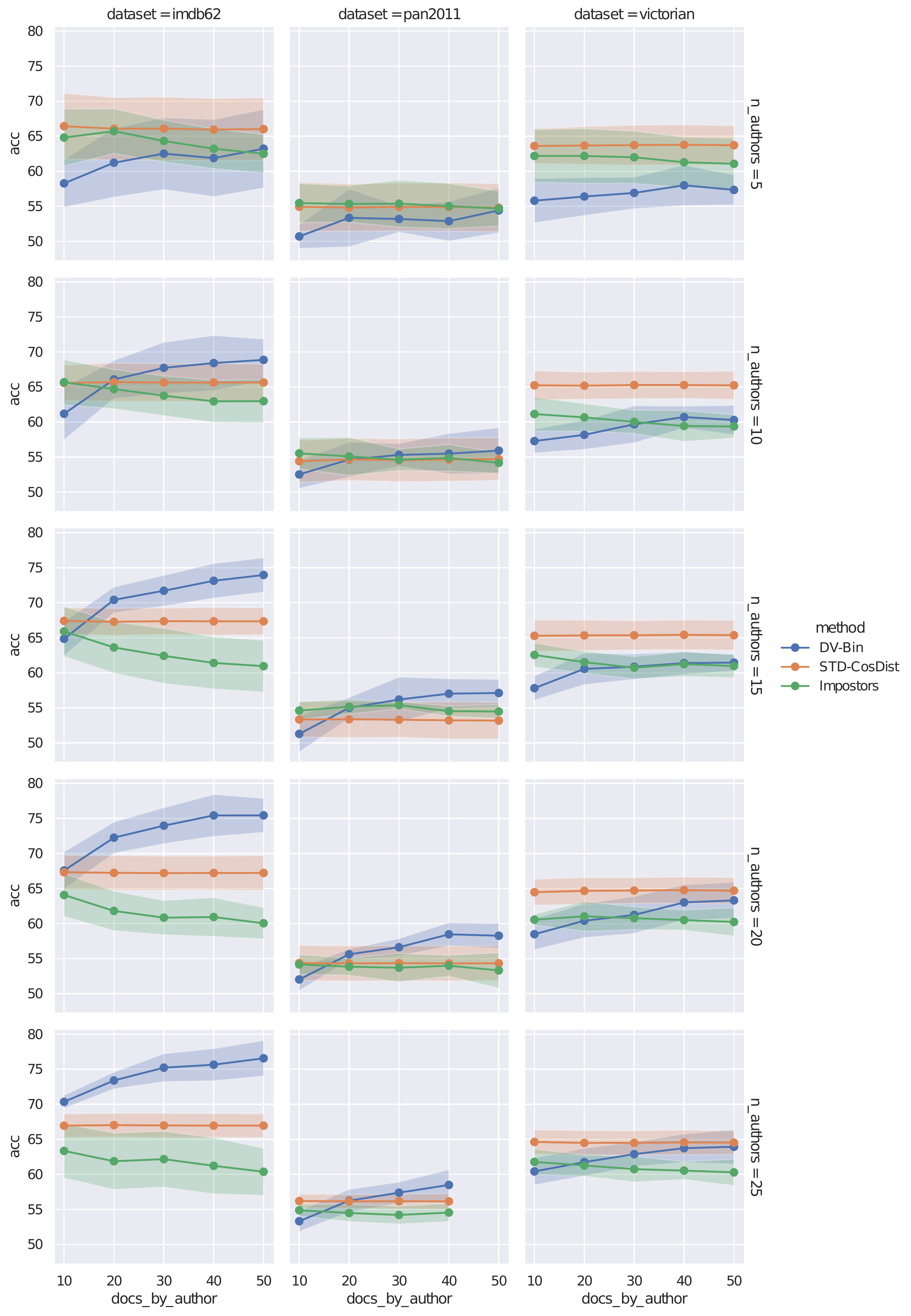}
    \caption{Intrinsic evaluation of \dv: results on open-set SAV,
    using vanilla accuracy (on the $y$ axis) as the evaluation measure
    on datasets \texttt{IMDB62}, \texttt{PAN2011}, and
    \texttt{Victorian}.}
    \label{fig:acc-open}
  \end{center}
\end{figure}

\noindent Figure \ref{fig:acc-open} displays the experimental results
we have obtained on \texttt{IMDB62}, \texttt{PAN2011}, and
\texttt{Victorian}, while Table \ref{tab:sav:arxiv:open} reports the
results obtained for the \texttt{arXiv} dataset.

\begin{table}[t]
  \begin{tabular}{|l|rrl|}
    \hline
    & \multicolumn{1}{c}{mean} & \multicolumn{1}{c}{std} & 
                                                           \multicolumn{1}{c|}{ttest} \\
    \hline
    DV-Bin & \textbf{.663} & 1.966 & \\
    STD-CosDist & .661 & 1.891 & ** \\
    Impostors & .642 & 2.473 & ** \\

    \hline
  \end{tabular}
  \caption{Intrinsic evaluation of \dv: results on open-set SAV, using
  vanilla accuracy as the evaluation measure on dataset
  \texttt{arXiv}. The notational conventions are the same as in
  Table~\ref{tab:sav:arxiv:close}.}
  \label{tab:sav:arxiv:open}
\end{table}
There is no clear winner in the light of these results. DV-Bin seems
to perform best in \texttt{IMDB62}, especially when the number of
authors increases; all methods seem to perform comparably in
\texttt{PAN2011} and \texttt{arXiv}, and STD-CosDist seems to perform
slightly better in \texttt{Victorian}. Somehow surprisingly, the
Impostors method seems not to take advantage of the increase in the
number of documents per author, likely because the number of actual
impostors ($i=10$) is set in advance and thus the method is
indifferent to variations in $q$. DV-Bin tends to perform poorly when
the number of documents per author is very small (i.e., 10); this may
be explained by the fact that the number of \textsf{Same} pairs that
can be generated from 10 elements is relatively small. Concerning
STD-CosDist, it proves a fairly stable method, as in the closed-set
scenario. 
\texttt{PAN2011} proves the hardest dataset here, with all methods
performing only marginally better than a random classifier (which
would obtain an expected accuracy of 0.50). Regarding the
\texttt{arXiv} dataset, DV-Bin 
performs best on average, but the t-test reveals that this superiority
is not significant from a statistical point of view.

Summing up, there is no strong enough empirical evidence to claim that
the DV-Bin method outperforms Impostors in open-set SAV. However,
there are some technical reasons why one should prefer the DV-Bin
method to the Impostors method. The first concerns its
efficiency. Impostors is a lazy method, meaning that it has no offline
training phase, i.e., all inductive inference is carried out in the
classification phase, and the workload that a single test pair entails
is significant, since it involves computing the similarity between the
test document and each training document, and computing $k$ rounds of
bagging for each impostor and for each element in the
pair. Conversely, once trained, classifying an unlabelled pair using
\diffvectors\ comes down to computing a simple linear combination of
feature differences.\footnote{Of course, it is fair to mention that
the Impostors method incurs no cost for training. But this only
applies if the value of the parameter $\sigma^*$ is hard-wired. In
practice, the optimal $\sigma^*$ has to be estimated in a validation
phase, which amounts to using a training set to perform repeated
rounds of tests which, as indicated above, require a considerable
computational effort.} The second reason concerns its
applicability. By definition, the Impostors method cannot be used, as
observed above, in closed-set SAV and, more generally, in SAV settings
in which documents written by any of the authors of the test pair are
observed in training. The reason is that the method would likely
consider training documents by one of the test authors as candidate
impostors (since these training documents are expected to be more
similar to the test document), and thus the test author could wrongly
be taken for an impostor of herself.

Figure \ref{fig:scores:open} shows the distribution of the decision
scores (i.e., the posteriors $\Pr(\textsf{Same}|x',x'')$) for
\textsf{Same} pairs and \textsf{Different} pairs that Impostors and
\diffvectors\ compute. As for closed-set SAV, we set $m=20$ and $q=50$
for all datasets except \texttt{arXiv}, for which we instead set
$m=50$ and keep all documents per author. Recall that, in our open-set
setting, $m$ specifies both the number of authors involved in the
training set \emph{and} the number of authors involved in test (e.g.,
in the case of \texttt{arXiv}, we are using the entire dataset since
there are 100 distinct authors). For ease of visualisation, we report
the score values according to a logarithmic scale.

\def \firstcolfile {Impostors-open-log} \def \firstmethod {Impostors}
\def \secondcolfile{DiffVectors_open} \def \secondmethod {DV-Bin} \def
\captionlabel {Distribution of decision scores for positive and
negative (i.e., \textsf{Same} and \textsf{Different}) pairs as
computed by the Impostors method (1st column; note the log scale) and
by the DV-Bin method (2nd column).}  \def \plotlabel{fig:scores:open}

The Impostors method produces decision scores which tend to be very
close to 0. The dashed vertical line indicates the decision threshold
found optimal in the validation phase; this threshold is
$\sigma^*=0.005$
in all cases but in \texttt{arXiv}, where $\sigma^*=0.01$ worked
better. Note that this threshold succeeds in placing most of the
negative scores below it, but still misclassifies many
positives. Particularly, in \texttt{PAN2011} and \texttt{arXiv} it
fails to push many of the positive scores beyond the decision
threshold.

The DV-Bin method instead succeeds at polarising the decision scores
of \textsf{Same} and \textsf{Different} pairs in \texttt{IMDB62} and
\texttt{arXiv}, although it fails to allocate most of the negative
mass below the 0.5 threshold in \texttt{Victorian} and, to a greater
extent, in \texttt{PAN2011}.

Overall, as clear from a simple visual inspection, the DV-Bin method
is better than the Impostors method at correctly separating the scores
of the \textsf{Same} pairs from those of the \textsf{Different} pairs
on each of our four datasets.


\subsection{Extrinsic evaluation of \diffvectors}
\label{sec:extrinsic}

\noindent Our ``extrinsic'' evaluation of DVs consists of closed-set
AA experiments. We do not run experiments for AV since, as discussed
in Section~\ref{sec:methodav}, each of our AA experiments is also a
set of $m$ AV experiments, and can be evaluated as such.


\subsubsection{The AA results}
\label{sec:AAresults}

\noindent At the core of our AA methods is a SAV classifier that
operates on pairs of documents. Given a test document $x$, attribution
for it is performed by applying a combination rule to the posterior
probabilities generated for pairs of documents consisting of the test
document $x$ and a training document $x'$. In particular, we explore:

\begin{itemize}
\item Lazy AA: the lazy combination rule inspired by $k$-NN discussed
  in Section~\ref{sec:knn}.
\item Stacked AA: the linear combination rule inspired by stacked
  generalisation discussed in Section~\ref{sec:stacking}.
\end{itemize}

\noindent In these experiments we consider a set of authors
$\mathcal{A}=\{A_{1}, \ldots, A_{m}\}$, with $m>2$, each one having
$q$ training documents. Given a test set $\mathcal{U}$, the method is
asked to attribute each test document to one of the authors in
$\mathcal{A}$, in a single-label multiclass fashion.

We investigate the impact on AA accuracy of the number $m$ of authors
and the number $q$ of documents per author. We let $m$ take values in
the set $\{5,10,15,20,25\}$ as before, and we let $q$ take values in
the set $\{5,10, ..., 45,50\}$.\footnote{The reason why we explore a
finer-grain grid for $q$ with respect to our previously discussed SAV
experiments is that in this case we are not considering Impostors as a
competitor, and thus these experiments are considerably faster to
run. Note also that, differently from our SAV experiments, we here
report experiments also for the combination (\texttt{PAN2011},25,50)
since here we are only considering the closed-set setting, and since
in \texttt{PAN2011} there are at least 25 authors with 50 documents.}
As in Section \ref{sec:intrinsic}, and for analogous reasons, the
experiments are different for the \texttt{arXiv} dataset, in which the
above fine-grained exploration is not possible. For both Lazy AA and
Stacked AA, we use DV-Bin as the underlying SAV mechanism.

In this case, instead of vanilla accuracy we use $F_{1}$ as the
evaluation measure, since not all our datasets are balanced, and since
vanilla accuracy is, differently from $F_{1}$, a notoriously bad
measure for working with imbalanced datasets.
For all datasets we report the values of macro-averaged $F_{1}$, i.e.,
$F_{1}$ averaged across the authors in $\mathcal{A}$; in the case of
the \texttt{arXiv} dataset, we also report micro-averaged $F_{1}$
(i.e., $F_{1}$ as obtained on a global contingency table generated by
all the classification predictions for all authors), since this is the
only imbalanced dataset of the lot (and since micro-averages would
coincide with macro-averages in the perfectly balanced datasets
\texttt{IMDB62}, \texttt{PAN2011}, and \texttt{Victorian}). All
results are reported as averages across 10 runs that use different
random seeds.

As our baseline we consider STD-AA, a single-label multiclass
classifier trained to distinguish among the $m$ classes from the
observation of ``standard'' vectors of features. Given a test
document, the classifier returns the author which obtains the maximum
posterior probability.
%
%

\begin{figure}[t]
  \begin{center}
    \includegraphics[width=.85\textwidth]{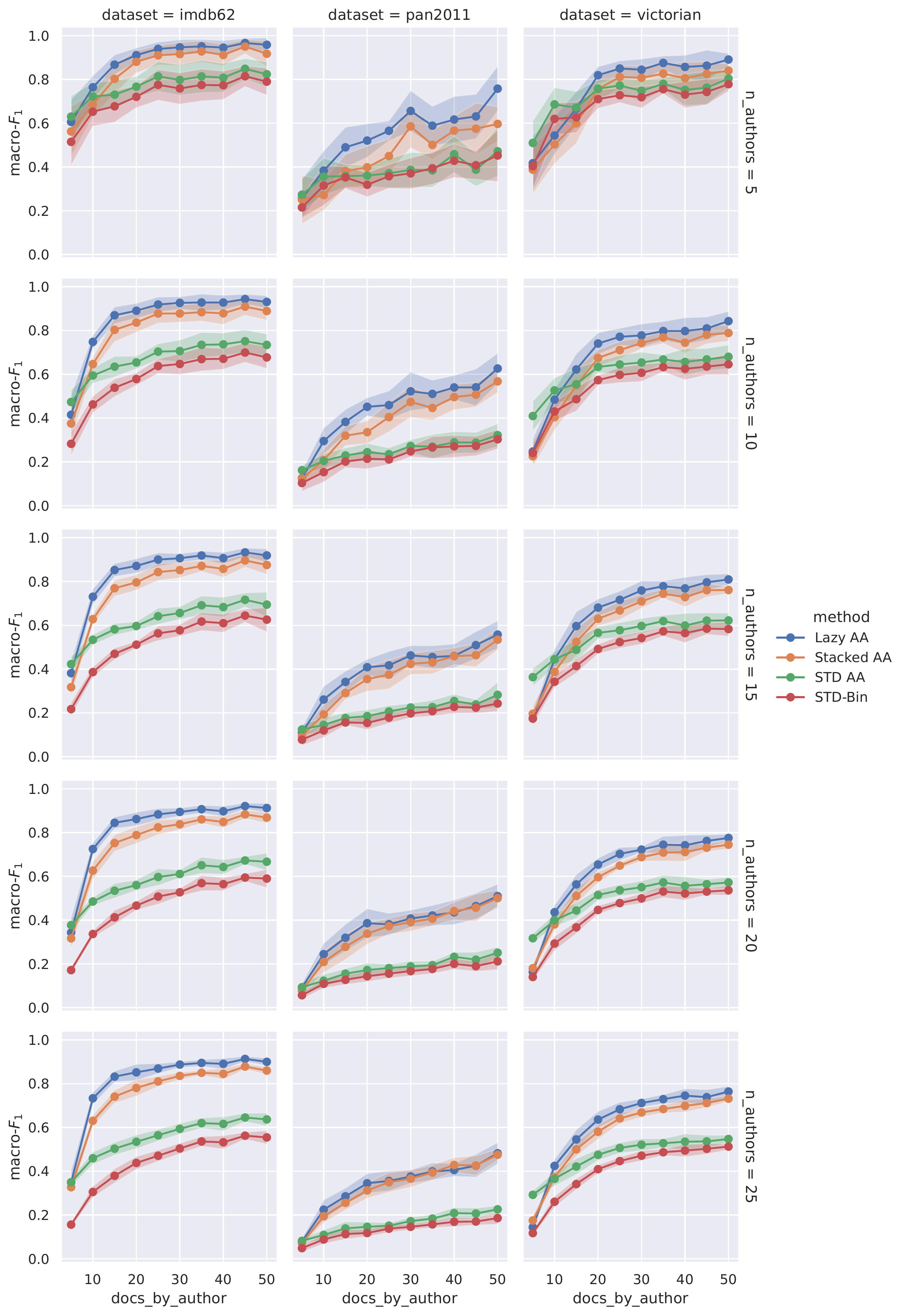}
 \caption{Extrinsic evaluation of \dv: results on closed-set AA in
 terms of $F_{1}$ for the \texttt{IMDB62}, \texttt{PAN2011}, and
 \texttt{Victorian} datasets. 
 }
 \label{fig:f1-attr}
\end{center}
\end{figure}

Figure \ref{fig:f1-attr} displays the experimental results we have
obtained for the \texttt{IMDB62}, \texttt{PAN2011}, and
\texttt{Victorian} datasets (for the moment being let us disregard the
curves for STD-Bin, on which we will comment later), while the first
three rows of Table \ref{tab:aa:arxiv:close} report the results
obtained on the \texttt{arXiv} dataset.

\begin{table}[t]
  \begin{tabular}{|l|rrl|rrl|}
    \hline
    & 
      \multicolumn{3}{c|}{macro-$F_{1}$} & 
                                           \multicolumn{3}{c|}{micro-$F_{1}$} \\
    \multicolumn{1}{|c|}{Method} & 
                                   \multicolumn{1}{c}{mean} & 
                                                              \multicolumn{1}{c}{std} & 
                                                                                        \multicolumn{1}{c|}{ttest} & 
                                                                                                                     \multicolumn{1}{c}{mean} & 
                                                                                                                                                \multicolumn{1}{c}{std} & 
                                                                                                                                                                          \multicolumn{1}{c|}{ttest} \\

    \hline
    Lazy AA & \textbf{0.643} & 0.244 & & \textbf{0.679} & 0.224 & \\
    Stacked AA & 0.596 & 0.237 & & 0.629 & 0.219 & \\
    STD-AA & 0.482 & 0.214 & & 0.508 & 0.207 & \\
    STD-Bin & 0.422 & 0.211 & & --- & --- & \\
 
    \hline
  \end{tabular}
  \caption{Extrinsic evaluation of \dv: results on closed-set AA in
  terms of $F_{1}$ for the \texttt{arXiv} dataset. The notational
  conventions are the same as for Table \ref{tab:sav:arxiv:close}.}
  \label{tab:aa:arxiv:close}
\end{table}

These results show the drastic superiority of DVs over standard
vectors for AA. Only for $q=5$, and infrequently for $q=10$, does STD
achieve (marginally) better results than the \dv-based variants; in
this case, the reason might have to do with the fact that low values
of $q$ result in fewer \textsf{Same} pairs (e.g., for $q=5$ there are
only 10 unordered pairs), which might lead to subobptimal accuracy for
the underlying SAV methods. Regarding our variants, the $k$-NN
-inspired combination rule consistently outperforms the linear one in
\texttt{IMDB62} and \texttt{arXiv}, and is slightly better or
comparable in the rest of the cases. All methods understandably
benefit from the increase in $q$, but DVs seem to do so at a much
greater rate; indeed, the increase in the number of training examples
is quadratic in $q$ for the \dv-based variants, while it is linear in
$q$ for STD.


\subsubsection{The AV results}
\label{sec:AVresults}

\noindent Concerning the AV task, note that macro-averaged $F_{1}$ is
also the right measure for evaluating AV; in fact, $F_{1}$ as measured
on a specific author $A^{*}$ is the right measure for evaluating AV
once $A^{*}$ is considered the candidate author, and macro-averaged
$F_{1}$ is the right measure for computing the average performance for
all possible choices of $A^{*}$. As a consequence, the results
reported in Figure \ref{fig:f1-attr} and Table
\ref{tab:aa:arxiv:close} also count as an evaluation of the reported
methods for the AV task.

For AV, we add another baseline (which we call STD-Bin), which
consists of a binary classifier trained to distinguish between $A^{*}$
and $\overline{A^{*}}$ from the observation of ``standard'' vectors of
features; it is fair to add this baseline since it would be just
natural to solve AV by means of a binary classifier, instead of by
means of a multiclass classifier as STD-AA does.

However, the experimental results show STD-Bin to be inferior to
STD-AA, as clear from both Figure \ref{fig:f1-attr} and Table
\ref{tab:aa:arxiv:close}. This is in keeping with the results of our
preliminary experiments (discussed in Section~\ref{sec:methodav}) that
had convinced us to abandon the idea of performing AV via Lazy AV and
Stacked AV, in favour of versions of Lazy AA and Stacked AA in which
we attribute document $x$ to $A^{*}$ if the AA algorithm does so and
we attribute $x$ to $\overline{A^{*}}$ if the AA algorithm attributes
it to an author $A_{z}$ different from $A$. Concerning the likely
reasons why this happens, the same considerations we made in
Section~\ref{sec:methodav} apply.

In sum, given that STD-Bin is not a serious contender, the same
considerations on the superiority of DV-based methods over standard
methods that we had made in Section~\ref{sec:AAresults} for AA also
apply to AV.






 


\subsection{Efficiency}
\label{sec:efficiency}

\noindent The improvements in performance obtained by DV-based methods
with respect to methods based on standard vectorial representations
can be attributed to the increase in the number of training examples
resulting from pairing documents. However, this can be expected to
come at a computational cost. In this section we compare the actual
cost of DV-based methods with that of methods based on standard
representations.


\subsubsection{Efficiency analysis}

\noindent Let $n=|\mathcal{L}|$ be the number of training
documents. Let also assume that the total cost of training a
classifier is bounded by some function $f$ on the number $n$ of
training documents, a cost which depends on the learning algorithm and
its implementation; in other words, this total cost is $O(f(n))$,
where we can safely assume $f(n)$ to grow faster than or equal to
$n$. (We take the number of features as constant, which means that
this number does not impact our analysis of efficiency.) Let also
assume that the classification of a document requires constant time,
i.e., is $O(1)$.

The cost of training the DV-Bin classifier of
Section~\ref{sec:methodsav} comes down to the cost of generating the
$n(n-1)/2$ pairs, which is $O(n^2)$, plus the cost of training a
classifier using $n(n-1)/2$ DVs, which is $O(f(n^2))$. In practice,
and in order to keep the computational burden under reasonable bounds,
we only generate a fixed number of examples (i.e., we avoid generating
all pairs first and discarding some of them later). Let $n=mq$, with
$m$ the number of authors and $q$ the number of documents per author,
as before; we generate all $mq(q-1)/2$ pairs of type \textsf{Same} and
as many pairs of type \textsf{Different}, thus ending up with
$mq(q-1)$ documents, which has a cost $O(mq^2)=O(nq)$, plus, again,
the cost of training the classifier from the $mq(q-1)$ documents,
which is $O(f(nq))$; since we have assumed $f(n)$ to grow faster than
or equal to $n$, the total cost of generating a DV-Bin classifier is
$O(f(nq))$. At classification time we only need to compute the
absolute difference between two vectors and invoke the classifier; for
most classifiers (and for LR in particular) this cost can be
considered constant, i.e., $O(1)$.

As a lazy algorithm, Lazy AA does not involve any real training
phase. However, it seeks for the optimal value of $k$, and this
entails pre-computing a matrix of distances, which is done only once
(and is $O(n^2)$), plus sorting, for each of the $m$ authors and for
each of the $n$ training documents (see Equation~\ref{eq:dvknn}), the
$q$ training documents by this author (which is $O(q \log
q)$). Altogether, this entails a total cost of
$O(n^2 + mnq\log q)=O(n^2 \log q)$. At classification time (for both
the AA and the AV settings), we only need to sort, for each of the $m$
training authors, the $q$ training documents by this author, which
means that this is $O(mq\log q)=O(n\log q)$.

Concerning Stacked AA, training the system entails (i) training a
DV-Bin classifier, which, as argued above, has a cost $O(f(nq))$; (ii)
creating the projections $\phi(x)$ for each of the $n$ training
documents, which has a cost $O(n^{2})$ (since creating one such
projection has a cost $O(n)$); (iii) training the metaclassifier on
the $n$ vectors $\phi(x)$ thus generated, which has a cost $O(f(n))$;
the total cost of training the system is thus the larger of $O(n^{2})$
and $O(f(nq))$. At classification time we need to generate the
representation $\phi(x)$ of the test document, which has a cost
$O(n)$, and to invoke the meta-classifier, which we can assume to
require constant time.

The Impostors method does not properly carry out a training phase, but
incurs the cost of optimising the $\sigma$ parameter, which consists
of carrying out $t$ rounds of test, with $t$ a user-defined
parameter. The computational cost of testing whether two documents
have been written by the same author or not entails computing, for
each of the $n$ training instances, the similarity with each test
document, which is $O(n)$, plus sorting by similarity in order to
choose the ``impostors'', which is $O(n \log n)$; this means that the
total cost is $O(n \log n)$. Impostors then performs $k$ rounds of
bagging trials with respect to each of the $i$ impostors, which adds a
cost $O(ki)$ if we assume the similarity function to be computed in
constant time.

Table~\ref{tab:efficiency} summarizes all the costs involved.

\begin{table}[h]
  \center
  \begin{tabular}{cccc}
    \toprule
    & Tasks & Training & Test \\
    \midrule
 
    STD & AV, AA & $O(f(n))$ & $O(1)$ \\
 
    \dv & SAV & $O(f(nq))$ & $O(1)$ \\
    Lazy AA & AV, AA & $O(n^2\log q)$ & $O(n \log q)$ \\
 
    Stacked AA & AV, AA & $\max\{O(n^2), O(f(nq))\}$ & $O(n)$ \\
 
    Impostors & SAV & $O(n \log n)$ & $O(n \log n)$ \\
 
    \bottomrule
 
  \end{tabular}
  \caption{Computational cost of a number of algorithms discussed in
  this paper.}
  \label{tab:efficiency}
\end{table}


\subsubsection{Timings}

\noindent As for the experiments reported in Figures
\ref{fig:scores:close} and \ref{fig:scores:open}, we report actual
timings clocked for $m=20$ and $q=50$ in the case of the
\texttt{IMDB62}, \texttt{PAN2011}, and \texttt{Victorian} datasets,
and for the entire dataset in the case of \texttt{arXiv}.
The variables that influence the analysis include the number of
training documents ($|\mathcal{L}|$), the number of pairs generated by
\dv\ ($|\mathcal{L}_{\mathcal{P}}|$), and the number of test documents
($|\mathcal{U}|$). Recall that $|\mathcal{L}_{\mathcal{P}}|$ depends
on the number of \textsf{Same} pairs that can be generated, which is
fixed and amounts to $20(50\cdot 49)/2$=24,500 for \texttt{IMDB62},
\texttt{PAN2011}, and \texttt{Victorian}, and which is variable and
depends on the random split (we report the value averaged across 10
runs) for \texttt{arXiv}. The values are summarised in Table
\ref{tab:variables} for convenience. Recall that the number of test
pairs in SAV tasks is fixed for all datasets and is equal to
1,000. Note that the \texttt{arXiv} dataset is split differently for
SAV and AA since, although we used the entire dataset in both tasks,
in the former we held half the authors out for composing the open set.
All times refer to computations carried out on the same machine,
equipped with a 12-core processor Intel Core i7-4930K at 3.40GHz with
32 GB of RAM, under Ubuntu 18.04. All methods run on CPU and are
implemented using \texttt{scikit-learn} and the \texttt{SciPy}
stack. We have parallelised all parallelisable steps, both in training
and test, for all algorithms.

\begin{table}[t]
  \center
  \begin{tabular}{|l|rrr|}
    \hline
    & \multicolumn{1}{c}{$|\mathcal{L}|$} & \multicolumn{1}{c}{$|\mathcal{L}_{\mathcal{P}}|$} & 
                                                                                                \multicolumn{1}{c|}{$|\mathcal{U}|$} \\
    \hline
    \texttt{IMDB62} & 1,000 & 49,000 & 6,000 \\
    \texttt{PAN2011} & 1,000 & 49,000 & 463 \\
    \texttt{Victorian} & 1,000 & 49,000 & 7,937 \\
    \texttt{arXiv}-SAV & 518 & 5,784 & 255 \\
    \texttt{arXiv}-AA & 1,028 & 11,106 & 441 \\
    \hline
  \end{tabular}
  \caption{Size of the datasets used for the efficiency test.}
  \label{tab:variables}
\end{table}


Table \ref{timings:sav} reports the average time each method requires
to complete the SAV task, both in terms of training time and testing
time for each dataset. The method that uses standard vectors to
compute the cosine distance (STD-CosDist) is much faster than any
competing method, both in terms of training times and test times. This
is due to the fact that cosine can be computed very quickly, and that
the classifier operates on one single feature. At training time, both
DV-Bin and Impostors are computationally much more expensive, with
neither one being clearly better than the other. However, at
classification time DV-Bin is much faster than Impostors, and costs no
more than a few seconds to accomplish the 1,000 SAV computations,
comparably to STD-CosDist. Impostors, on the contrary, requires much
more time, and its testing times are higher than its training
times. (Recall that, for the Impostors method, by ``training'' we mean
the search for the optimal value of parameter $\sigma$ by using the
training set, since Impostors does not properly perform any training.)

\begin{table}[t]
  \center
  \begin{tabular}{|l|rr|rr|rr|rr|}
    \hline
    & 
      \multicolumn{2}{c|}{\texttt{IMDB62}} & 
                                             \multicolumn{2}{c|}{\texttt{PAN2011}} & 
                                                                                     \multicolumn{2}{c|}{\texttt{Victorian}} & 
                                                                                                                               \multicolumn{2}{c|}{\texttt{arXiv}-SAV} \\
    & 
      \multicolumn{1}{c}{Train} & 
                                  \multicolumn{1}{c|}{Test} & 
                                                              \multicolumn{1}{c}{Train} & 
                                                                                          \multicolumn{1}{c|}{Test} & 
                                                                                                                      \multicolumn{1}{c}{Train} & 
                                                                                                                                                  \multicolumn{1}{c|}{Test} & 
                                                                                                                                                                              \multicolumn{1}{c}{Train} & 
                                                                                                                                                                                                          \multicolumn{1}{c|}{Test} \\ \hline
    DV-Bin & {\ul 438.3} & 1.4 & 173.7 & \textbf{0.7} & 
                                                        {\ul 870.2} & 2.5 & 49.8 & 0.3 \\
    STD-CosDist & \textbf{6.7} & \textbf{0.3} & \textbf{3.5} & 
                                                               0.8 & \textbf{11.9} & \textbf{0.3} & \textbf{0.7} & 
                                                                                                                   \textbf{0.2} \\
    Impostors & 271.9 & {\ul 625.1} & {\ul 247.6} & 
                                                    {\ul 455.7} & 283.2 & {\ul 651.3} & {\ul 225.2} & 
                                                                                                      {\ul 224.6} \\ \hline
  \end{tabular}
  \caption{Training and testing times (in seconds) clocked when
  solving the SAV task. \textbf{Boldface} and {\ul underlining}
  indicate the fastest and slowest methods for each dataset,
  respectively.}
  \label{timings:sav}
\end{table}

Table \ref{timings:attr} reports the average time each method requires
to complete the AA Task.
It is immediately evident that, at least on \texttt{IMDB62},
\texttt{PAN2011} and \texttt{Victorian}, the two most expensive
methods are the ones based on DVs, both at training time and at
classification time (neither one is systematically better or worse
than the other, though); the STD method is thus almost always the
fastest. The reason for this high computational cost of the DV-based
methods is that, despite the fact that DV-Bin proved very fast at
classification time in SAV, Lazy AA and Stacked AA invoke DV-Bin
\emph{many} times, i.e., require computing, for all training documents
(in the training phase) and for all test documents (in the testing
phase), the similarity (viewed as a posterior probability computed by
DV-Bin) with each training document. This has an important impact both
in the training phase and in the testing phase.

Although the increase in training time with respect to the SAV task is
not marked, the penalty paid during the testing phase is instead
evident; for \texttt{arXiv}, training times of the DV-based methods
increase substantially with respect to those seen for the SAV task,
which is due to the fact that in this case the training set is twice
as large as that for SAV -- see Table \ref{tab:variables}). In some
cases (\texttt{IMDB62} and \texttt{Victorian}) testing times even
surpass training times; this can be explained by the fact that those
datasets contain the largest tests sets (6,000 and 7,937 instances,
respectively), which means that computing the matrix of posterior
probabilities becomes especially costly.

Somehow surprisingly, though, the variants based on DV-Bin were
trained faster than STD in \texttt{arXiv}; the reason for this lies in
the number of authors involved, which in this dataset is the largest,
i.e., $m=100$.
STD thus needs to train 100 binary classifiers on a
document-by-feature matrix of $O(10^{5})$ dimensions, while DV-based
variants need to train only one binary classifier in order to discern
between \textsf{Same} or \textsf{Different}; note also that in this
case the number of pairs generated is comparatively smaller than for
other datasets. The rest of the work that DV-based methods undertake
is on a matrix of posterior probabilities that has just
$|\mathcal{L}|$=1,028 dimensions in the case of \texttt{arXiv};
training 100 binary classifiers in Stacked AA is thus much faster than
with STD.

To conclude, DVs bring about substantially higher computational costs
than the ``standard'' representations, both at training time and at
test time. However, we should note that in most typical authorship
analysis endeavours the additional computational costs are usually not
a matter of concern, if compensated by increases in accuracy (as they
are here).

Concerning training times, the increases that DV-based methods bring
about are tolerable, since training is carried out once for all, and
the times reported in Tables~\ref{timings:sav} and~\ref{timings:attr}
are plausible for most application contexts; note also that, in most
authorship analysis applications, training documents are often scarce,
which means that scenarios in which the training documents are many
more than in our datasets (which would mean training times even higher
than those reported in Tables~\ref{timings:sav}
and~\ref{timings:attr}) are unfortunately infrequent.

Concerning classification times, we argue that increased costs are
usually tolerable in real authorship identification cases, because
typical such cases \emph{do not involve many unlabelled
documents}. Indeed, there is often a \emph{single} unlabelled
document, of extremely high value, that we need to make a prediction
for (e.g., a text of literary value~\cite{Benedetto:2013mb,
Corbara:2019cq, Mosteller:1964gb, Savoy:2019qr, Tuccinardi:2017yg}, or
an anonymous letter), and in this case issuing a prediction in
milliseconds or in minutes does not make a big difference.

\begin{table}[t]
  \center
  \begin{tabular}{|l|rr|rr|rr|rr|}
    \hline
    & \multicolumn{2}{c|}{\texttt{IMDB62}} & \multicolumn{2}{c|}{\texttt{PAN2011}} & \multicolumn{2}{c|}{\texttt{Victorian}} & \multicolumn{2}{c|}{\texttt{arXiv}-AA} \\
    & \multicolumn{1}{c}{Train} & \multicolumn{1}{c|}{Test} & \multicolumn{1}{c}{Train} & \multicolumn{1}{c|}{Test} & \multicolumn{1}{c}{Train} & \multicolumn{1}{c|}{Test} & \multicolumn{1}{c}{Train} & \multicolumn{1}{c|}{Test} \\ \hline
    Lazy AA & 460.2 & {\ul 955.5} & 228.3 & {\ul 61.8} & 917.4 & 1232.6 & \textbf{145.0} & 66.1 \\
    Stacked AA & {\ul 480.3} & 942.8 & {\ul 267.3} & 61.2 & {\ul 955.4} & {\ul 1235.9} & 250.4 & {\ul 66.5} \\
    STD-AA & \textbf{156.0} & \textbf{0.2} & \textbf{157.6} & \textbf{0.1} & \textbf{202.9} & \textbf{0.5} & \underline{483.8} & \textbf{0.1} \\
    \hline
  \end{tabular}
  \caption{Training and testing times (in seconds) clocked for solving
  the AA task. \textbf{Boldface} and {\ul underlining} indicate the
  fastest and slowest methods for each dataset, respectively.}
  \label{timings:attr}
\end{table}







%
%
%
%
%
%


\subsection{Can we use \diffvectors\ for ``natively binary'' AV
problems?}
\label{sec:nativeav}

\noindent So far, we have tested DVs in situations in which, at
training time, we assume we know who among the $n$ authors in
$\mathcal{A}$ has written which training documents. That is, we have
recast SAV, AA, and AV in terms of a multiclass task. In this section
we turn to analyse experimentally the suitability of DVs for AV in a
different situation, i.e., one in which all we know about a certain
training document is whether it has been written by the author of
interest or not, that is, whether this document is a positive example
or a negative example with respect to a binary classification scheme.

To this aim, we randomly draw $m=10$ authors for each dataset, and
perform, for each author, an AV experiment in which we take this
author as the positive class and the rest of the authors (grouped
together) as the negative class, and where we employ an AA method
(i.e., one that was originally devised for tackling arbitrary values
of $n$) for the particular case of $n=2$. That is, given
$\mathcal{A}=\{A_{1}, A_{2}, \ldots, A_{10}\}$, we generate a binary
setting $\mathcal{A}'_i=\{A_{i}, \overline{A}_{i}\}$, and we do this
for all authors by letting $i$ vary in the range $\{1,\ldots,10\}$. In
each of these experiments, we take $q=50$ documents for each author in
$\mathcal{A}=\{A_{1}, A_{2}, \ldots, A_{10}\}$ in all datasets but in
\texttt{arXiv}, for which we take all the documents available for the
author. We repeat the entire process 10 times with different random
seeds and report results averaged across all experiments. The results
reported in Table~\ref{tab:nativeAV} show that, in such a setting, DVs
do not bring about any benefit.

\begin{table}[ht!]
  \centering \resizebox{\textwidth}{!}{
  \begin{tabular}{|l|rrr|rrr|rrr|rrr|}\hline
    & \multicolumn{3}{c|}{\texttt{IMDb62}} & \multicolumn{3}{c|}{\texttt{PAN2011}} & \multicolumn{3}{c|}{\texttt{Victorian}} & \multicolumn{3}{c|}{\texttt{arXiv}} \\
    \hline
    & mean & std & ttest & mean & std & ttest & mean & std & ttest & mean & std & ttest \\
    \hline
    Lazy AA & 0.287 & 0.044 & & 0.166 & 0.019 & & 0.205 & 0.010 & & 0.234 & 0.037 & \\
    Stacked AA & 0.664 & 0.062 & ** & 0.278 & 0.056 & ** & 0.619 & 0.044 & ** & 0.365 & 0.075 & ** \\
    STD-Bin & \textbf{0.683} & 0.045 & & \textbf{0.285} & 0.028 & & \textbf{0.639} & 0.042 & & \textbf{0.432} & 0.097 & \\

    \hline
  \end{tabular}
  }
  \caption{Results, in terms of macro-$F_1$, obtained by applying AA
  methods to ``native'' AV problems.}
  \label{tab:nativeAV}
\end{table}

This was somehow to be expected, since DVs bring useful additional
evidence to the learning process for AV only when we have access to
the entire labelling information, i.e., when author $A_j$ can help
improve classification for author $A_i$ indirectly, by strengthening
the internal SAV function with additional instances of the class
\textsc{Same} that come from documents written by $A_j$ (see
Section~\ref{sec:moretrainingdata}). This is not possible in a pure
binary setting, since the negative class \emph{is not homogeneous}
(i.e., it does not represent the production of one single author, but
the production of many authors that have been mixed together), and
thus cannot be leveraged to generate positive instances for the class
\textsc{Same}. For similar reasons, we cannot generate instances of
\textsc{Different} by simply picking two documents from the negative
class, since those could have been written by the same (unknown)
author. What we are left with, thus, is the possibility to generate
$q(q-1)/2$ instances of \textsc{Same} only from pairs of instances
from the positive class $A_i$, and $q^2(m-1)$ instances of
\textsc{Different} by generating pairs in which one document has been
written by $A_{i}$ and the other by $\overline{A}_{i}$. Since the
positive evidence for the surrogate SAV problem (class \textsc{Same})
comes exclusively from the positive class of the AV problem (author
$A_i$), there is no real information gain with respect to using
standard representations. Indeed, we observe a degradation in
performance of both variants with respect to the adoption of
``standard'' vector representations; this degradation is not
statistically significant for the stacking variant, though.

For this reason, we conclude that, in AV settings, DV-based methods
should be used only in situations in which we have access to the
entire class label information. Luckily enough, access to the entire
class label information is something that characterises most scenarios
in which AV is to be applied since, when investigating whether
document $x$ is indeed by author $A^{*}$ or not, it makes sense to
generate a training dataset in which negative instances are known to
be by authors ``close'' (in a stylistic sense) to $A^{*}$, and we can
know this only by knowing who the author of each document is.

\section{Related work}
\label{sec:related}

\noindent In the authorship analysis literature, example works
(starting from \citet{Koppel:2014bq}) in which two or more documents
are represented by a single vector have been presented before; the
main difference between those papers and the present one is that none
among the former performed any systematic study, as we instead do, of
the implications of the use of these representations. In the next
paragraphs we summarise the major approaches along this line.

As previously mentioned, \cite{Koppel:2014bq} was the first work in
which vectors each representing more than one document were used in
the authorship analysis literature. In this representation, a vector
represented two documents, the label of the vector was either
\textsf{Same} or \textsf{Different}, character 4-grams were used as
features, and the value of each feature was the absolute difference
between the tf-idf weights of the feature in the two documents. As
mentioned in the introduction, the goal of \citet{Koppel:2014bq} was
to propose a different method (the ``impostors'' method for SAV), and
they dismiss the DV-based representation as a ``simplistic baseline
method'' \cite[p.\ 179]{Koppel:2014bq}.

Since then, a number of authors started to view the AV task in terms
of predicting whether vector $f(X_{A^{*}},x)$ belongs to class
\textsf{Same} or to class \textsf{Different}, where $f(X_{A^{*}},x)$
is a vector derived from the entire set (here represented as
$X_{A^{*}}$) of training documents known to be by candidate author
$A^{*}$, and from the document of unknown paternity (here noted as
$x$). For instance, in \citep{Bartoli:2015tf} vector $f(X_{A^{*}},x)$
is a vector in which each feature value is the absolute difference
between the value of the feature in $x$ and the mean of the values of
the feature across the documents in $X_A$.
A slightly different approach is used in the PRNN method presented by
\citet{Hosseinia:2018ex}. They view $X_{A^{*}}$ as a document
(generated by the concatenation of all the documents in it), use both
this document and document $x$ as input for a parallel neural network
composed of an embedding layer and an RNN layer, and combine the two
outputs by computing a vector $f(X_{A^{*}},x)$ consisting of values of
similarity between the two documents. The same work also proposes a
different method, called TE, which is based on a transformation
encoder that transforms the vector representing $A^{*}$ into the
vector representing $x$, and takes the resulting loss as a measure of
similarity; the authors repeat the process several times using
different feature sets, and generate a vector $f(X_{A^{*}},x)$
consisting of the different similarity values. In a similar vein, in
\citep{Bevendorff:2019pl} the $f(X_{A^{*}},x)$ vector is composed of 7
similarity values computed on the char $n$-grams of the two
documents. Unlike the present work, none of the above works attempts
to tackle the AV and AA tasks by recasting them in terms of SAV.

More recently, \cite{Menta:2021by, Weerasinghe:2021fr, Ikae:2021ne}
tested the use of \dv\ for the open-set SAV problem at the recent
PAN2021 shared task; in particular, \citet{Menta:2021by} propose a
method that feeds DVs to a double-channel neural network, where the
feature values are the tf-idf weights of character $n$-grams in one
channel, and of punctuation marks in the other channel. The outputs of
the two channels are then concatenated in a final series of layers,
that ultimately leads to the classification decision.

Finally, we note that the DV-based representations that we have
discussed are reminiscent of ideas that have been independently
explored in multilingual text classification. In particular,
\citet{Moreo:2016fk} investigate the idea of applying lightweight
random projections to the feature space. Mathematically, a random
projection $X R$ of a matrix $X\in\mathbb{R}^{np}$, with $n$ the
number of documents and $p$ the number of features, can be attained by
multiplying it with a random matrix $R\in\mathbb{R}^{pr}$, with
$r\ll p$ the number of dimensions. The term ``lightweight'' refers to
the fact that the rows in $R$ contain only two non-zero values
(-1,+1). The pair-based version $\mathcal{L}_{\mathcal{P}}$ of a
dataset $\mathcal{L}$ can be defined in terms of $|R\cdot X|$, where
$X\in\mathbb{R}^{np}$ is our document-by-feature matrix and $R$ is
instead a lightweight projection matrix $R^{rn}$, this time with $r$,
the number of pairs, much higher than $n$; here $|\cdot|$ represents
the element-wise absolute value. Such a projection effectively
computes the absolute difference between two chosen documents.

\section{Conclusion}
\label{sec:conclusion}

\noindent In this paper we have discussed the implications of the use
of \diffvectors\ (DVs) in authorship identification tasks. A DV is a
vector that represents a pair of documents in such a way that the
value of a feature in the DV is the absolute difference between the
relative frequencies (or increasing functions thereof) of the feature
in the two documents. DVs were originally introduced by
\citet{Koppel:2014bq}, but in that very same work these authors
dismissed DVs as a ``simplistic baseline method''. Neither
\citet{Koppel:2014bq} nor other authors studied the implications of
the use of DVs in authorship identification; a systematic study of
these implications is what this paper describes.

DVs are naturally geared towards solving the ``same-author
verification'' (SAV) task, i.e., the binary task of deciding whether
two documents have been written by the \textsf{Same} (possibly
unknown) author or by \textsf{Different} authors. However, we have
shown that both (i) (closed-set) authorship attribution (the task of
predicting who among a given set of candidates is the true author of a
given text), and (ii) authorship verification (the task of predicting
whether a given author is or not the author of a given text), can be
recast in terms of SAV; we have presented two original algorithms
(\emph{Lazy AA} and \emph{Stacked AA}) that do this for both AA and
AV.

In order to compare DV-based authorship identification methods with
their counterparts based on ``standard'' vectors, we have carried out
experiments on four datasets of texts labelled by author (one of which
we have created ourselves and we here make publicly available for the
first time) and representative of different textual genres, lengths,
and styles, and on three authorship identification tasks (SAV, AA,
AV). Our experiments have shown that DV-based methods are particularly
suited to some authorship identification tasks and are not suited to
others. For instance, the results indicate that neither standard
methods nor DV-based methods clearly outperform each other on open-set
SAV (see Section~\ref{sec:sav:open}). Instead, DV-based methods vastly
outperform the competition on three important tasks, i.e., (a) on
closed-set SAV (see Section~\ref{sec:sav:close}), (b) on closed-set AA
(see Section~\ref{sec:extrinsic}), and (c) on AV (see
Section~\ref{sec:extrinsic}). As we have argued, these benefits derive
from the fact that, in many cases, DV-based methods may exploit more
training data than methods based on standard vectors (see
Section~\ref{sec:moretrainingdata}), and that DVs may make training
more robust also when the above is not the case (see
Section~\ref{sec:morerobusttraining}).


%
%
%
%
%
%

In future work we would like to study ``diff-functions'' other than
the absolute difference of (a static, fixed increasing function of)
the feature frequencies of the two documents, by testing the
possibility of dynamically \emph{learning} such functions from data,
in the style of \citep{Moreo:2020cq}.
Other aspects worth exploring include testing DVs in authorship
profiling tasks, such as native language identification.



\begin{acks}
  The authors' work has been supported by the \textsc{SoBigData++}
  project, funded by the European Commission (Grant 871042) under the
  H2020 Programme INFRAIA-2019-1, by the \textsc{AI4Media} project,
  funded by the European Commission (Grant 951911) under the H2020
  Programme ICT-48-2020, and by the \textsc{SoBigData.it},
  \textsc{FAIR} and \textsc{ITSERR} projects funded by the Italian
  Ministry of University and Research under the NextGenerationEU
  program. These authors' opinions do not necessarily reflect those of
  the funding agencies.
\end{acks}




\end{document}